\documentclass[10pt,twocolumn,letterpaper]{article}

\usepackage{cvpr}              %

\usepackage{graphicx}
\usepackage{amsmath}
\usepackage{amssymb}
\usepackage{booktabs}
\usepackage{subcaption}
\usepackage{capt-of}
\usepackage{color}
\usepackage{float}
\usepackage{array}
\usepackage{subfiles}
\usepackage{comment}
\usepackage[export]{adjustbox}

\usepackage[pagebackref,breaklinks,colorlinks]{hyperref}

\usepackage[capitalize]{cleveref}
\crefname{section}{Sec.}{Secs.}
\Crefname{section}{Section}{Sections}
\Crefname{table}{Table}{Tables}
\crefname{table}{Tab.}{Tabs.}

\def\cvprPaperID{6649} %
\def\confName{CVPR}
\def\confYear{2022}

\definecolor{orange}{rgb}{1.0,0.65,0.0}
\definecolor{mygreen}{rgb}{0.05,0.5,0.35}
\newif\ifcomments
\ifcomments
    \newcommand{\roy}[1]{\textcolor{blue}{[R: #1]}}
    \newcommand{\eli}[1]{\textcolor{mygreen}{[E: #1]}}
    \newcommand{\ira}[1]{\textcolor{green}{[I: #1]}}
    \newcommand{\jj}[1]{\textcolor{cyan}{[JJ: #1]}}
    \newcommand{\xuan}[1]{\textcolor{orange}{[X: #1]}}
    \newcommand{\mengyi}[1]{\textcolor{magenta}{[M: #1]}}
    
\else
    \providecommand{\roy}[1]{}
    \providecommand{\eli}[1]{}
    \providecommand{\ira}[1]{}
    \providecommand{\jj}[1]{}
    \providecommand{\xuan}[1]{}
    \providecommand{\mengyi}[1]{}
\fi

\newcommand\nomarkerfootnote[1]{%
  \begingroup
  \renewcommand\thefootnote{}\footnote{#1}%
  \addtocounter{footnote}{-1}%
  \endgroup
}

\newif\ifarxiv
\arxivtrue %

\begin{document}

\title{StyleSDF: High-Resolution 3D-Consistent Image and Geometry Generation}

\author{Roy Or-El\textsuperscript{1} \qquad
        Xuan Luo\textsuperscript{1} \qquad
        Mengyi Shan\textsuperscript{1} \qquad
        Eli Shechtman\textsuperscript{2} \\
        Jeong Joon Park\textsuperscript{3} \qquad
        Ira Kemelmacher-Shlizerman\textsuperscript{1} \\
        \textsuperscript{1}University of Washington \qquad
        \textsuperscript{2}Adobe Research \qquad
        \textsuperscript{3}Stanford University}

\makeatletter
\let\@oldmaketitle\@maketitle
\renewcommand{\@maketitle}{\@oldmaketitle%
\vspace{-0.5cm}
\centering
\includegraphics[width=1.0\linewidth]{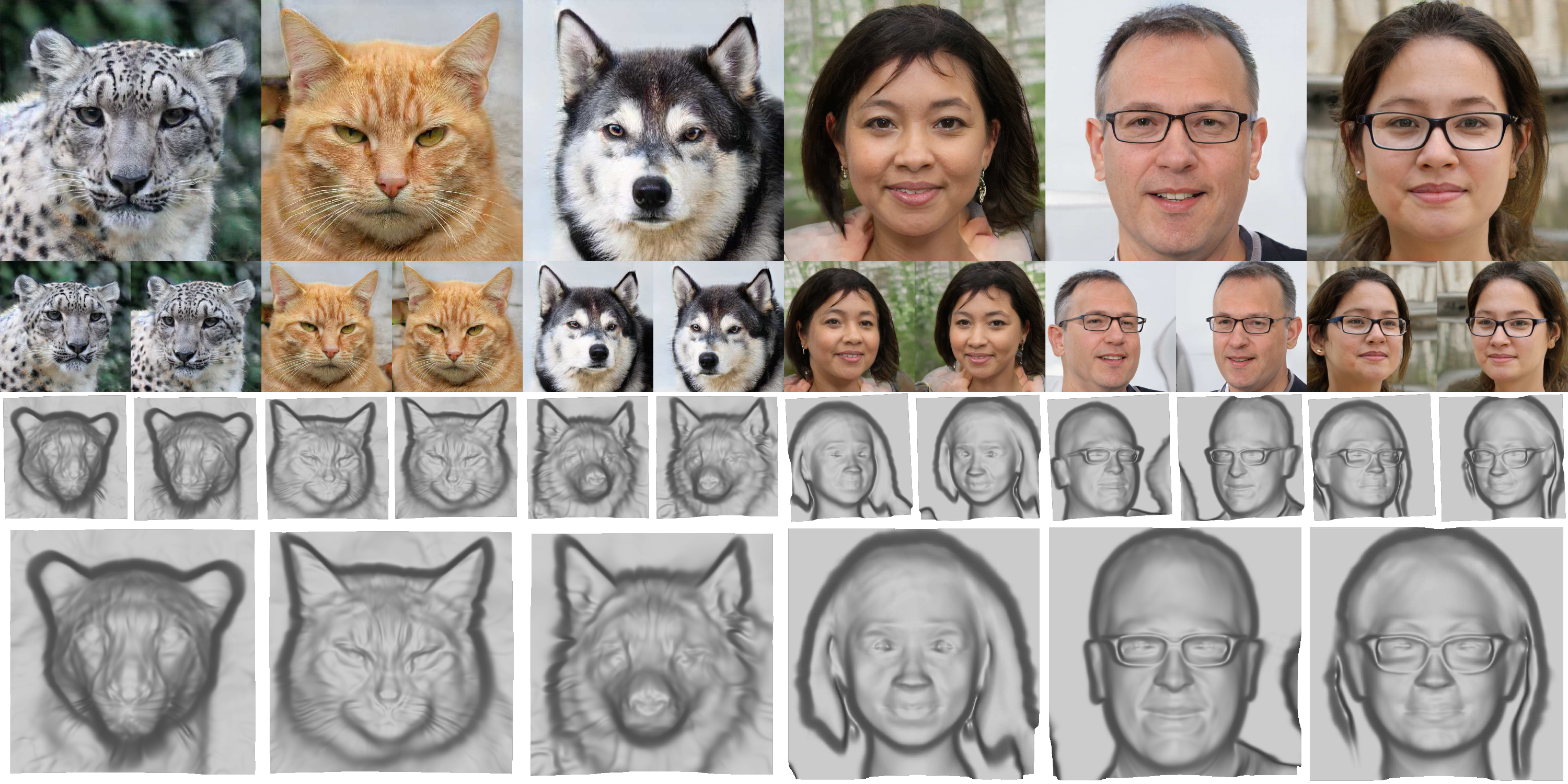}
\captionof{figure}{Our proposed framework--StyleSDF-- learns to jointly generate high resolution, 3D-consistent images (top rows) along with their detailed view-consistent geometry  represented with SDFs (depth maps in bottom rows), while being trained on single view RGB images.}
\label{fig:teaser}
\bigskip}
\makeatother
\maketitle

\nomarkerfootnote{Project Page: \url{https://stylesdf.github.io/}}
\begin{abstract}
We introduce a high resolution, 3D-consistent image and shape generation technique which we call StyleSDF. Our method is trained on  single-view RGB data only, and stands on the shoulders of StyleGAN2 for image generation, while solving two main challenges in 3D-aware GANs: 1) high-resolution, view-consistent generation of the RGB images, and 2) detailed 3D shape. We achieve this by merging a SDF-based 3D representation with a style-based 2D generator. Our 3D implicit network renders low-resolution feature maps, from which the style-based network generates view-consistent, 1024$\times$1024 images. Notably, our SDF-based 3D modeling defines detailed 3D surfaces, leading to  consistent volume rendering. Our method shows higher quality results compared to state of the art in terms of visual and geometric quality.
\end{abstract}

\vspace{-0.5cm}

\section{Introduction}
\label{sec:intro}

StyleGAN architectures \cite{karras2019style,karras2020analyzing, karras2021alias} have shown an unprecedented quality of RGB image generation. They are, however, designed to generate single RGB views rather than 3D content. In this paper, we introduce StyleSDF, a method for generating 3D-consistent 1024$\times$1024 RGB images and geometry, trained only on single-view RGB images.

Related 3D generative models \cite{park2019deepsdf,sitzmann2019scene, chan2021pi,niemeyer2021giraffe,Schwarz2020NEURIPS} present shape and appearance synthesis via coordinate-based multi-layer-perceptrons (MLP). These works, however, often require 3D or multi-view data for supervision, which are difficult to collect, or are limited to low-resolution rendering outputs as they rely on expensive volumetric field sampling. Without multi-view supervision, 3D-aware GANs \cite{Schwarz2020NEURIPS,niemeyer2021giraffe,chan2021pi} typically use opacity fields as geometric proxy, forgoing well-defined surfaces, which results in low-quality depth maps that are inconsistent across views.  

At the core of our architecture lies the SDF-based 3D volume renderer and the 2D StyleGAN generator. We use a coordinate-based MLP to model Signed Distance Fields (SDF) and radiance fields which render low resolution feature maps. These feature maps are then efficiently transformed into high-resolution images using the StyleGAN generator. Our model is trained with an adversarial loss that encourages the networks to generate realistic images from all sampled viewpoints, and an Eikonal loss that ensures proper SDF modeling. These losses automatically induce view-consistent, detailed 3D scenes, without 3D or multi-view supervision. The proposed framework effectively addresses the resolution and the view-inconsistency issues of existing 3D-aware GAN approaches that base on volume rendering. Our system design opens the door for interesting future research in vision and graphics that involves a latent space of high quality shape and appearance. 

Our approach is evaluated on the FFHQ~\cite{karras2019style} and AFHQ~\cite{choi2020stargan} datasets. We demonstrate through extensive experiments that our system outperforms the state-of-the-art 3D-aware methods, measured by the quality of the generated images and surfaces, and their view-consistencies.

\vspace{-0.2cm}
\section{Related Work}
\label{sec:related}
In this section, we review related approaches in 2D image synthesis, 3D generative modeling, and 3D-aware image synthesis.

\vspace{0.1cm}\noindent\textbf{Generative Adversarial Networks:}
State-of-the-art Generative Adversarial Networks~\cite{goodfellow2014generative} (GANs) can synthesize high-resolution RGB images that are practically indistinguishable from real images~\cite{karras2017progressive,karras2019style,karras2020analyzing,karras2021alias}. 
Substantial work has been done in order to manipulate the generated images, by exploring meaningful latent space directions~\cite{abdal2019image2stylegan,tewari2020stylerig,shen2020interpreting,jahanian2019steerability,abdal2020image2stylegan++,collins2020editing,tewari2020pie,harkonen2020ganspace,shen2021closed,abdal2021styleflow}, introducing contrastive learning~\cite{Shoshan_2021_ICCV}, inverse graphics~\cite{zhang2020image}, examplar images~\cite{kafri2021stylefusion} or multiple input views~\cite{leimkuhler2021freestylegan}. While 2D latent space manipulation produces realistic results, these methods tend to lack explicit camera control, have no 3D understanding, require shape priors from 3DMM models~\cite{tewari2020stylerig,tewari2020pie}, or reconstruct the surface as a preprocessing step~\cite{leimkuhler2021freestylegan}.

\vspace{0.1cm}\noindent\textbf{Coordinate-based 3D Models:} 
While multiple 3D representations have been proposed for generative modeling \cite{yang2019pointflow,wu2016learning,groueix2018papier}, recent coordinate-based neural implicit models \cite{park2019deepsdf, mescheder2019occupancy,chen2019learning} stand out as an efficient, expressive, and differentiable representation. 

Neural implicit representations (NIR) have been widely adopted for learning shape and appearance of objects~\cite{michalkiewicz2019implicit,niemeyer2019occupancy,saito2019pifu,saito2020pifuhd,chibane20ifnet,atzmon2020sal,davies2020overfit,icml2020_2086, oechsle2019texture}, local  parts~\cite{genova2019learning,genova2020local}, and full 3D scenes~\cite{chabra2020deep,chibane2020ndf,peng2020convolutional,jiang2020local} from explicit 3D supervisions. 
Moreover, NIR approaches have been shown to be a powerful tool for reconstructing 3D structure from multi-view 2D supervision via fitting their 3D models to the multi-view images using differentiable rendering \cite{sitzmann2019scene,yariv2020multiview,niemeyer2020differentiable,mildenhall2020nerf}.

Two recent seminal breakthroughs are NeRF~\cite{mildenhall2020nerf} and SIREN~\cite{sitzmann2020implicit}. NeRF introduced the use of volume rendering~\cite{kajiya1984ray} for reconstructing a 3D scene as a combination of neural radiance and density fields to synthesize novel views. SIREN replaced the popular ReLU activation function with sine functions with modulated frequencies, showing great single scene fitting results. We refer readers to ~\cite{tewari2021advances} for more comprehensive review.

\vspace{0.1cm}\noindent\textbf{Single-View Supervised 3D-Aware GANs:} 
Rather than relying on 3D or multi-view supervisions, recent approaches aim at learning a 3D generative model from a set of unconstrained single-view images. These methods ~\cite{Schwarz2020NEURIPS,nguyen2019hologan,chan2021pi,niemeyer2021giraffe,niemeyer2021campari,BlockGAN2020,lunz2020inverse,henderson2020leveraging,gadelha20173d,jimenez2016unsupervised} typically optimize their 3D representations to render realistic 2D images from all randomly sampled viewpoints using adversarial loss.

\begin{figure*}[t!]
    \centering
    \includegraphics[width=1.0\linewidth]{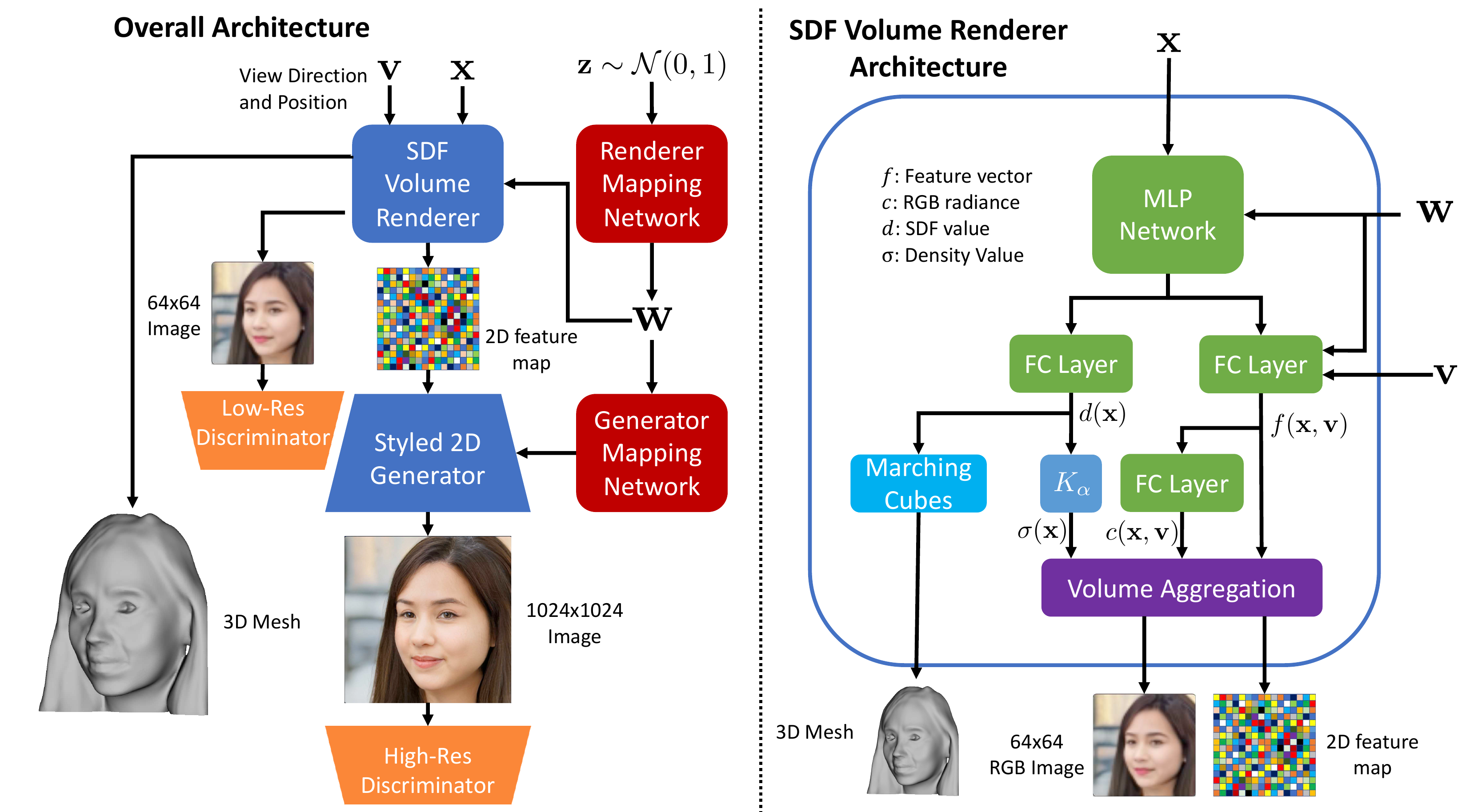}
    \vspace{-0.65cm}
    \caption{StyleSDF Architecture: (Left)  Overall architecture: SDF volume renderer takes in a latent code and camera parameters, queries points and view directions in the volume, and projects the 3D surface features into the 2D view. The projected features are fed to the Styled 2D generator that creates the high resolution image. (Right) our SDF volume renderer jointly models volumetric SDF and radiance field, providing a well defined and view consistent geometry.}
    \label{fig:overview}
    \vspace{-0.6cm}
\end{figure*}

Most inline with our work are methods that use 
implicit neural radiance fields for 3D-aware image and geometry generation (GRAF~\cite{Schwarz2020NEURIPS} and Pi-GAN~\cite{chan2021pi}).
However, these methods are limited to low-resolution outputs due to the high computational costs of the volume rendering.
In addition, the use of density fields as proxy for geometry provides ample amount of leeway for the networks to produce realistic images while violating 3D consistency, leading to inconsistent volume rendering \wrt the camera viewpoints (the rendered RGB or depth images are not 3D-consistent).

To improve the surface quality, ShadeGAN~\cite{pan2021shadegan} introduces a shading-guided pipeline, 
and GOF~\cite{xu2021generative} gradually shrink the sampling region of each camera ray. 
However, the image output resolution (128$\times$128) is still bounded by the computational burden of the volume rendering.
 GIRAFFE~\cite{niemeyer2021giraffe} proposed a dual stage rendering process. A backbone volume renderer generates low resolution feature maps (16$\times$16) that are passed to a 2D CNN to generate outputs at $256\times256$ resolution. 
 Despite improved image quality, GIRAFFE outputs lack view consistency. The hairstyle, facial expression, and sometimes the object's identity, are entangled with the camera viewpoint inputs, likely because 3D outputs at 16$\times$16 are not descriptive enough.

 Concurrent works~\cite{gu2022stylenerf,zhou2021cips,Chan2022,deng2022gram} adopt two-stage rendering process or smart sampling procedures for high-resolution image generation, yet these works still do not model well-defined, view-consistent 3D geometry.

\section{Algorithm}
\label{sec:algo}
\subsection{Overview}
Our framework consists of two main components. A backbone conditional SDF volume renderer, and a 2D style-based generator~\cite{karras2020analyzing}. Each component also has an accompanied mapping network~\cite{karras2019style} to map the input latent vector into modulation signals for each layer.  An overview of our architecture can be seen in~\figurename~\ref{fig:overview}.

To generate an image, we sample a latent vector $\mathbf{z}$ from the unit normal distribution, and camera azimuth and elevation angles $(\phi, \theta)$ from the dataset's estimated object pose distribution. For simplicity, we assume that the camera is positioned on the unit sphere and directed towards the origin. Next, our volume renderer outputs the signed distance value, RGB color, and a 256 element feature vector for all the sampled volume points along the camera rays.
We calculate the surface density for each sampled point from its SDF value and apply volume rendering \cite{mildenhall2020nerf} to project the 3D surface features into 2D feature map. The 2D generator then takes the feature map and generates the output image from the desired viewpoint. The 3D surface can be visualized with volume-rendered depths or with the mesh from marching-cubes algorithm~\cite{lorensen1987marching}.

\subsection{SDF-based Volume Rendering}
Our backbone volume renderer takes a 3D query point, $\mathbf{x}$ and a viewing direction $\mathbf{v}$. Conditioned by the latent vector $\mathbf{z}$, it outputs an SDF value $d(\mathbf{x},\mathbf{z})$, a view dependent color value $\mathbf{c}(\mathbf{x},\mathbf{v},\mathbf{z})$, and feature vector $\mathbf{f}(\mathbf{x},\mathbf{v},\mathbf{z})$. For clarity, we omit $\mathbf{z}$ from hereon forward.

The SDF value indicates the distance of the queried point from the surface boundary, and the sign indicates whether the point is inside or outside of a watertight surface. As shown in VolSDF~\cite{yariv2021volume}, the SDF can be serve as a proxy for the density function used for the traditional volume rendering \cite{mildenhall2020nerf}. Assuming a non-hollow surface, we convert the SDF value into the 3D density fields $\sigma$, 
\begin{equation}
    \sigma(\mathbf{x}) = K_{\alpha} \left( d(x) \right) = \frac{1}{\alpha} \cdot \text{Sigmoid}\left(\frac{-d(\mathbf{x})}{\alpha}\right),
\end{equation}
where  $\alpha$ is a learned parameter that controls the tightness of the density around the surface boundary. $\alpha$ values that approach $0$ represent a solid, sharp, object boundary, whereas larger $\alpha$ values indicate a more ``fluffy" object boundary. A large positive SDF value would drive the sigmoid function towards $0$, meaning no density outside of the surface, and a high-magnitude negative SDF value would push the sigmoid towards $1$, which means maximal density inside the surface. 

We render low resolution $64\times64$ feature maps and color images with volume rendering. For each pixel, we query points on a ray that originates at the camera position $\mathbf{o}$, and points at the camera direction $\mathbf{r}(t) = \mathbf{o} + t\mathbf{v}$. and calculate the RGB color and feature map as follows:
\begin{equation}
    \begin{aligned}
         &\mathbf{C}(\mathbf{r}) = \int_{t_n}^{t_f}T(t)\sigma(\mathbf{r}(t))\mathbf{c}(\mathbf{r}(t),\mathbf{v})dt,\\
         &\mathbf{F}(\mathbf{r}) = \int_{t_n}^{t_f}T(t)\sigma(\mathbf{r}(t))\mathbf{f}(\mathbf{r}(t),\mathbf{v})dt,\\
         &\text{where} \quad T(t) = \exp \left(-\int_{t_n}^t\sigma(\mathbf{r}(s))ds\right),
    \end{aligned}
\end{equation}
which we approximate with discrete sampling along rays.

Unlike NeRF\cite{mildenhall2020nerf} and other 3D-aware GANs such as Pi-GAN~\cite{chan2021pi} and StyleNeRF~\cite{gu2022stylenerf} we do not use stratified sampling. Instead, we split $[t_n,t_f]$ into $N$ evenly-sized bins, draw a single offset term uniformly $\delta \sim \mathcal{U}[0, \frac{t_f-t_n}{N}]$, and sample N evenly-spaced points,
\begin{equation}
    t_i = \frac{t_f-t_n}{N}\cdot i + \delta, \quad \text{where} \quad i \in \{0,\ldots,N-1\}.
\end{equation}
In addition, we forgo hierarchical sampling altogether, thereby reducing the number of samples by 50\%. We discuss the merits of our sampling strategy in the supplementary material. %

The incorporation of SDFs provides clear definition of the surface, allowing us to extract the mesh via Marching Cubes~\cite{lorensen1987marching}. Moreover, the use of SDFs along with the related losses (Sec.~\ref{sec: vol_render}) leads to higher quality geometry in terms of expressiveness and view-consistency (as shown in Sec.~\ref{sec:volume_rendering}), even with a simplified volume sampling strategy.%

 The architecture of our volume renderer mostly matches that of Pi-GAN~\cite{chan2021pi}. The mapping network consists of a 3 layer MLP  with LeakyReLU activation and maps an input latent code $\mathbf{z}$ into $\mathbf{w}$ space and then generates frequecny modulation, $\gamma_i$, and phase shift, $\beta_i$, for each layer of the volume renderer. The volume rendering network contains eight shared modulated FC layers with SIREN~\cite{sitzmann2020implicit} activation: 
\begin{equation}
    \phi_i(x) = \sin \left(\gamma_i(W_i\cdot x + b_i)+\beta_i \right), \quad i \in \{0, \ldots, 7\}
\end{equation}
where $W_i$ and $b_i$ are the weight matrix and bias vector of the fully connected layers. The volume renderer then splits into two paths, the SDF path and the color path. The SDF path is implemented using a single FC layer denoted $\phi_d$. In the color path, the output of the last shared layer $\phi_7$ is concatenated with the view direction input and passed into one additional FiLM siren layer~\cite{chan2021pi} $\phi_f$ followed by a single FC layer $\phi_c$ that generates the color output. To summarize:
\begin{equation}
    \begin{aligned}
         &\sigma(\mathbf{x}) = K_{\alpha} \circ \phi_d \circ \phi_7 \circ \ldots \circ \phi_0(\mathbf{x}),\\
         &f(\mathbf{x},\mathbf{v}) = \phi_f (\phi_7 \circ \ldots \circ \phi_0(\mathbf{x}), \mathbf{v)}\\
         &c(\mathbf{x},\mathbf{v}) = \phi_c \circ \phi_f.
    \end{aligned}
\end{equation}

The output features of $\phi_f$ are passed to the 2D style-based generator, and the generated low resolution color image is fed to a discriminator for supervision. The discriminator is identical to the Pi-GAN~\cite{chan2021pi} discriminator.

We observed that using view-dependent color $\mathbf{c}(\mathbf{x},\mathbf{v})$ tends to make the networks overfit to biases in the dataset. For instance, people in FFHQ~\cite{karras2019style} tend to smile more when  facing the camera. This makes the facial expression change with the viewpoint although the geometry remains consistent. However, when we removed view-dependent color, the model did not converge. Therefore, to get view consistent images, we train our model with view dependent color, but fix the view direction $\mathbf{v}$ to the frontal view during inference.

\subsection{High-Resolution Image Generation}
 Unlike NeRF~\cite{mildenhall2020nerf}, where the reconstruction loss is computed individually for each ray, adversarial training needs a full image to be present. Therefore, scaling a pure volume renderer to high-resolution quickly becomes untractable, as we need to sample over $10^7$ queries to render a single 1024$\times$1024 image. As such, we seek to fuse a volume renderer with the StyleGAN2 network that has a proven capabilities of synthesizing high-resolution 2D images.

To combine the two architectures, we truncate the early layers of the StyleGAN2 generator up until the $64\times64$ layer and feed the generator with the $64\times64$ feature maps generated by the backbone volume renderer. In addition, we cut StyleGAN2's mapping network from eight layers to five layers, and feed it with the $\mathbf{w}$ latent code from the volume renderer's mapping network, instead of the original latent vector $\mathbf{z}$. The discriminator is left unchanged. 

This design choice allows us to enjoy the best of both worlds. The volume renderer learns the underline geometry, explicitly disentangles the object's pose from it's appearance, and enables full control of the camera position during inference. The StyleGAN2 generator upsamples the low resolution feature maps, adds high frequency details, and mimics complex light transport effects such as sub-surface scattering and inter-reflections that are difficult to model with the low-resolution volume renderer. 

\begin{figure*}[t!]
    \centering
    \includegraphics[width=1.0\linewidth]%
    {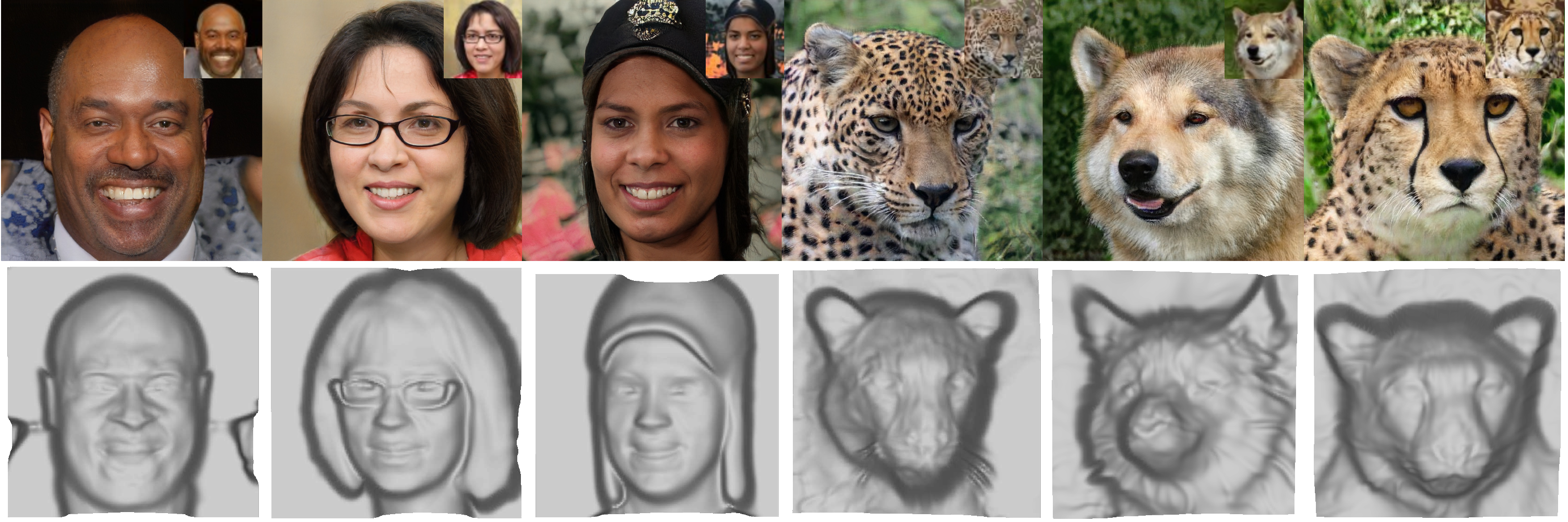}
    \caption{Generated high-res RGB images (top), low-res volume rendered images (inset) and depth maps (bottom) for the same view . The 64$\times$64 volume rendering output features are passed to the StyleGAN generator for high-resolution RGBs. Note that the object identities and structures are preserved between the image pairs. Furthermore, as can be seen in the jaguar and cheetah examples, the StyleGAN generator occasionally corrects badly modeled background signal from the volume renderer.}
    \label{fig:additional_results}
    \vspace{-0.4cm}
\end{figure*}

\subsection{Training}
We employ a two-stage training procedure. First we train only the SDF-based volume renderer, then we freeze the volume renderer weights, and train the StyleGAN generator.

\subsubsection{Volume Renderer training}\label{sec: vol_render}
We use the non-saturating GAN loss with R1 regularization~\cite{mescheder2018training}, denoted $\mathcal{L}_{adv}$, to train our volume renderer. On top of that, we use 3 additional regularization terms.

\noindent\textbf{Pose Alignment Loss:} This loss is designed to make sure that all the generated objects are globally aligned. %
On top of predicting whether the image is real or fake, the discriminator also tries to predict the two input camera angles ($\phi, \theta$). We penalize the prediction error using a smoothed L1 loss:
\begin{equation}
    \mathcal{L}_{view} = \begin{cases}
    (\hat{\theta} - \theta)^2   & \quad \text{if } |\hat{\theta} - \theta| \leq 1\\
    |\hat{\theta} - \theta|  & \quad \text{otherwise} 
  \end{cases}.
\end{equation}
This loss is applied on both view angles for the generator and the discriminator, however, since we don't have ground truth pose data for the original dataset, this loss is only applied to the fake images in the discriminator pass.

\noindent\textbf{Eikonal Loss:} This term ensures that the learned SDF is physically valid \cite{gropp2020implicit}:
\begin{equation}
    \mathcal{L}_{eik} = \mathbb{E}_{\mathbf{x}}(\|\nabla d(\mathbf{x})\|_2- 1)^2.
\end{equation}

\noindent\textbf{Minimal Surface Loss:} We encourage the 3D network to describe the scenes with minimal volume of zero-crossings to prevent spurious and non-visible surfaces from being formed within the scenes. That is, we penalize the SDF values that are close to zero:
\begin{equation}
    \mathcal{L}_{surf} = \mathbb{E}_{\mathbf{x}} \left( \exp(-100|d(x)|) \right ).
    \label{eq:min_surf}
\end{equation}

The overall loss function is then,
\begin{equation}
    \mathcal{L}_{vol} = \mathcal{L}_{adv} + \lambda_{view}\mathcal{L}_{view} + \lambda_{eik}\mathcal{L}_{eik} + \lambda_{surf}\mathcal{L}_{surf},
\end{equation}
where $\lambda_{view} = 15$, $\lambda_{eik} = 0.1$, and $\lambda_{surf} = 0.05$. The weight of the R1 loss is set according to the dataset.

\subsubsection{Styled Generator Training}
We train our Styled generator with the same losses and optimizer parameters as the original implementation, a non saturating adversarial loss, R1 regularization, and path regularization. As in the volume renderer training, we set the weight of the R1 regularization according to the dataset.

While it is possible to have a reconstruction loss between the low-resolution and high-resolution output images, we find that the inductive bias of the 2D convolutional architecture and the sharing of style codes is strong enough to preserve important structures and identities between the images (Fig. \ref{fig:additional_results}).

\begin{figure*} 
\includegraphics[width=1.0\linewidth]{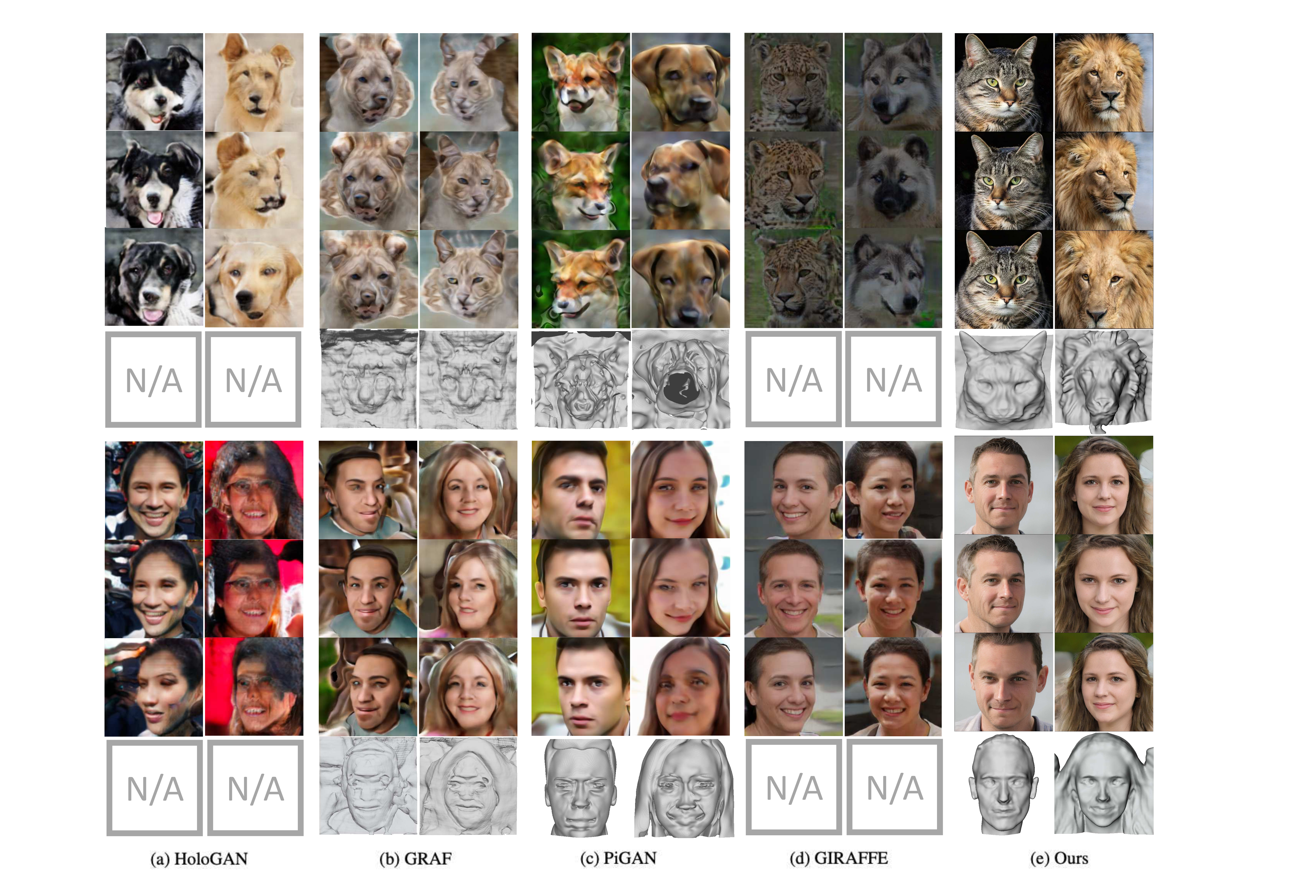}
\caption{Qualitative image and geometry comparisons. We compare our sample renderings and corresponding 3D meshes against the state-of-the-art 3D-aware GAN approaches (\cite{nguyen2019hologan,Schwarz2020NEURIPS,chan2021pi,niemeyer2021giraffe}). Note that HoloGAN and GIRAFFE are unable to create 3D mesh from their representations. Both HoloGAN (a) and GRAF (b) produce renderings that are of lower quality. The 3D mesh reconstructed from PiGAN's learned opacity fields reveal noticeable artifacts (c). While GIRAFFE (d) produces realistic low-resolution images, the identity of the person often changes with the viewpoints. StyleSDF (d) produces 1024$\times$1024 realistic view consistent RGB, while also generating high quality 3D. Best viewed digitally.}
\label{fig:qualitative_comp}
\vspace{-0.5cm}
\end{figure*}

\section{Experiments}
\label{sec:exp}

\subsection{Datasets \& Baselines}
We train and evaluate our model on the FFHQ~\cite{karras2019style} and AFHQ~\cite{choi2020stargan} datasets. FFHQ contains 70,000 images of diverse human faces at $1024 \times 1024$ resolution, which are centered and aligned according to the procedure introduced in Karras~\etal~\cite{karras2017progressive}. The AFHQ dataset consists of 15,630 images of cats, dogs and wild animals at $512 \times 512$ resolution. Note that the AFHQ images are not aligned and contain diverse animal species, posing a significant challenge to StyleSDF.

We compare our method against the state-of-the-art 3D-aware GAN baselines, GIRAFFE~\cite{niemeyer2021giraffe}, PiGAN~\cite{chan2021pi}, GRAF~\cite{Schwarz2020NEURIPS} and HoloGAN~\cite{nguyen2019hologan}, on the above datasets by measuring the quality of the generated images, shapes, and rendering consistency.

\subsection{Qualitative Evaluations}

\noindent\textbf{Comparison to Baseline Methods:} We compare the visual quality of our images to the baseline methods by rendering the same identity (latent code) from 4 different viewpoints, results are shown in~\Cref{fig:qualitative_comp}. To compare the quality of the underlying geometry, we also show the surfaces extracted by marching cubes from StyleSDF, Pi-GAN, and GRAF (Note that GIRRAFE and HoloGAN pipelines do not generate shapes). Our method generates superior images as well as more detailed 3D shapes. Additional generation results from our method can be seen in~\Cref{fig:teaser,fig:additional_results}.

\begin{figure}[t!]
    \centering
    \includegraphics[width=1.0\linewidth]{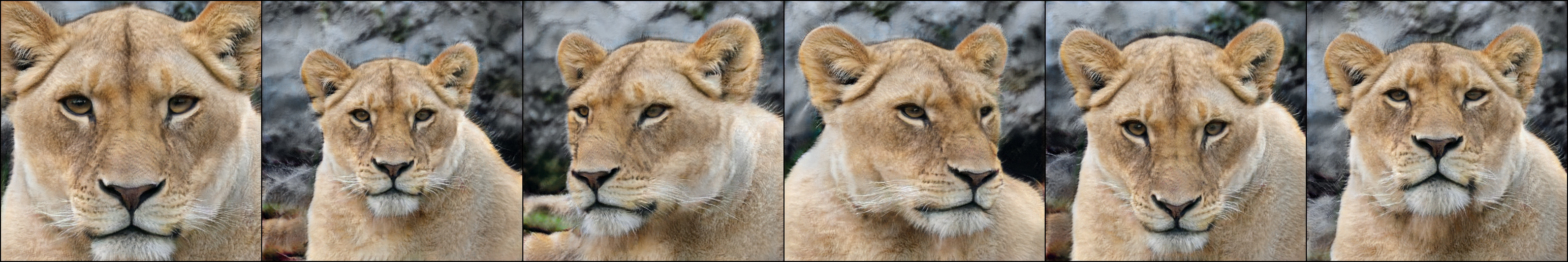} \\ 
    \includegraphics[width=1.0\linewidth]{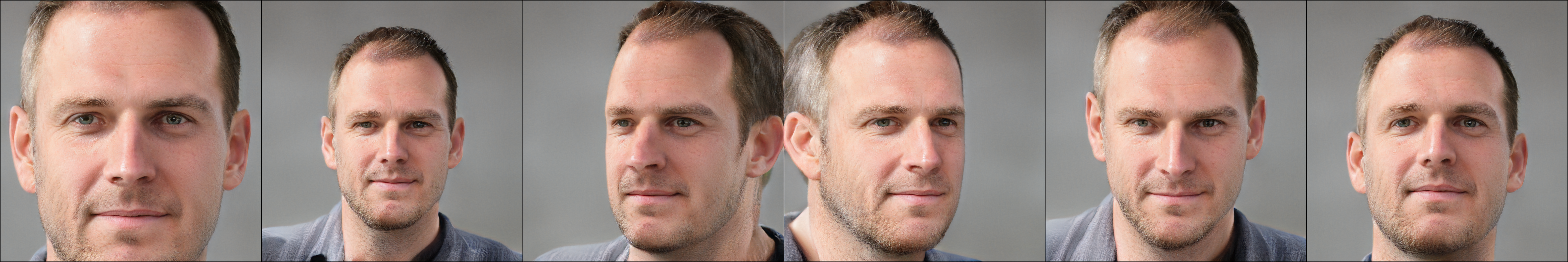}
    \caption{Out-of-distribution view synthesis (field of view and camera angles). Although StyleSDF was trained with a fixed field of view, increasing and decreasing FOV by $25\%$ (columns 1-2) still looks realistic. Similarly with 1.5 standard deviations of the camera angles distribution used for training (columns 3-6).}
    \label{fig:novel_views}
\end{figure}

\noindent\textbf{Novel View Synthesis:} Since our method learns strong 3D shape priors, it can generate images from viewpoints that are not well represented in the dataset distribution. Examples of out-of-distribution view synthesis are displayed in~\Cref{fig:novel_views}.

\noindent\textbf{Video Results:} We urge readers to view our \href{https://stylesdf.github.io/}{project's website} that includes a larger set of results and videos to better appreciate the multi-view capabilities of StyleSDF.\roy{change the reference from supplementary to project website}

\subsection{Quantitative Image Evaluations}
We evaluate the visual quality and the diversity of the generated images using the Frechet Inception Distance (FID)\cite{heusel2017fid} and Kernel Inception Distance (KID)\cite{binkowski2018demystifying}. We compare our scores against the aforementioned baseline models on the FFHQ and AFHQ datasets. 

All the baseline models are trained following their given pipelines to generate $256 \times 256$ images, with the exception of Pi-GAN, which is trained on $128 \times 128$ images and renders $256 \times 256$ images at inference time. The results, summarized in~\Cref{tbl:fid_kid}, show that StyleSDF performs consistently better than all the baselines in terms of visual quality. It is also on par with reported scores from concurrent works such as StyleNerf~\cite{gu2022stylenerf} and CIPS-3D~\cite{zhou2021cips}. 

\subsection{Volume Rendering Consistency}\label{sec:volume_rendering}
Volume rendering has emerged as an essential technique to differentiably optimize a volumetric field from 2D images, as its wide-coverage point sampling leads to stable gradient-flow during training. Notably, volume rendering excels at modeling thin surfaces or transparent objects, e.g., human hairs, which are difficult to model with explicit surfaces, e.g., 3D meshes. 

However, we notice that the volume rendering of existing 3D-aware GANs~\cite{chan2021pi,Schwarz2020NEURIPS} using unregularized opacity fields severely lacks view-consistency due to the absence of multi-view supervision.
That is, depth values, computed as the expected termination distance of each camera ray \cite{mildenhall2020nerf,deng2021depth}, from different viewpoints do not consistently overlap in the global coordinate. This means that neural implicit features are evaluated at inconsistent 3D locations, undermining the inductive bias of the implicit 3D representation for view-consistent renderings. As such, we measure and compare the depth map consistency across viewpoints to gauge the quality of volume rendering for each system.

\begin{table}[t]
\centering
\setlength{\tabcolsep}{12pt}
\begin{tabular}{lcccc}
\toprule
Dataset:     & \multicolumn{2}{c}{FFHQ} & \multicolumn{2}{c}{AFHQ} \\
\cmidrule(lr){2-3} \cmidrule(lr){4-5} 
             & FID  & KID              & FID  & KID \\
\midrule
HoloGAN      & 90.9   & 75.5               & 95.6   & 77.5  \\
GRAF         & 79.2   & 55.0               & 129.5   & 85.1  \\
PiGAN        & 83.0    & 85.8               & 52.4    & 30.7  \\
GIRAFFE      & 31.2    & 20.1               & 33.5    & 15.1  \\
\midrule
Ours         & \textbf{11.5} & \textbf{2.65}  & \textbf{12.8} & \textbf{4.47} \\
\bottomrule
\end{tabular}

\caption{FID and KID evaluations. All datasets were evaluated at a resolution of $256\times256$. Our method demonstrates the best performance. Note that we report KID $\times$ 1000 for simplicity.}
\label{tbl:fid_kid}
\vspace{-0.3cm}
\end{table}

\begin{table}[t]
\centering
\setlength{\tabcolsep}{20pt}
    \begin{tabular}{lrr}
    \toprule
    Dataset:     & FFHQ & AFHQ \\
    \midrule
    PiGAN        & 11.04    & 8.66    \\
    Ours         & \textbf{0.40}    & \textbf{0.63}    \\
    \bottomrule
    \end{tabular}

\caption{Depth consistency results. We measure the average modified Chamfer distance (\cref{eq:modified_chamfer_dist}) over 1,000 random pairs of depth maps for each dataset. Each pair contains one frontal view depth map and one side view depth map. Our method demonstrates significantly stronger consistency (see~\cref{fig:depth_consistency_viz}).}
\label{tbl:depth_consistency}
\end{table}

We sample 1,000 identities, render their $128\times 128$ depth maps from the frontal view and a fixed side view, and compute the alignment between the two views. The depth value is defined as the expected termination distance of $128$ uniformly sampled points along each ray. %
Note that we remove non-terminating rays whose accumulated opacity is below 0.5. We set the side viewpoint to be $1.5\times$ the standard deviation of the azimuth distribution in training. See supplementary for more experiment details.

To measure the alignment errors between the depth points, we adopt a modified Chamfer distance metric. I.e., 
we replace the usual mean distance definition with the median of the distances to nearest points,
\begin{equation}
    \begin{aligned}
        \mbox{CD}(S_1,S_2)=\underset{x\in S_1}{\mbox{med}}\min_{y\in S_2} \Vert x-y \Vert_2^2+ \underset{y\in S_2}{\mbox{med}}\min_{x\in S_1} \Vert x-y \Vert_2^2,
    \end{aligned}
    \label{eq:modified_chamfer_dist}
\end{equation}
for some point sets $S_1$ and $S_2$. This metric is more robust to outliers that come from occlusion and background mismatch that we are not interested in measuring. To put the metric at scale, we normalize the distances by the volume sampling bin size. %

\begin{figure}[t]
    \centering
    \setlength{\tabcolsep}{0pt}
    \def\imW{0.25\linewidth}
    \begin{tabular}{p{\imW}p{\imW}p{\imW}p{\imW}}
        \multicolumn{1}{c}{Ours} &  \multicolumn{1}{c}{PiGAN} & \multicolumn{1}{c}{Ours}&\multicolumn{1}{c}{PiGAN} \vspace{-1.1em}\\
        & & & \\
        \multicolumn{4}{c}{\includegraphics[width=\linewidth]{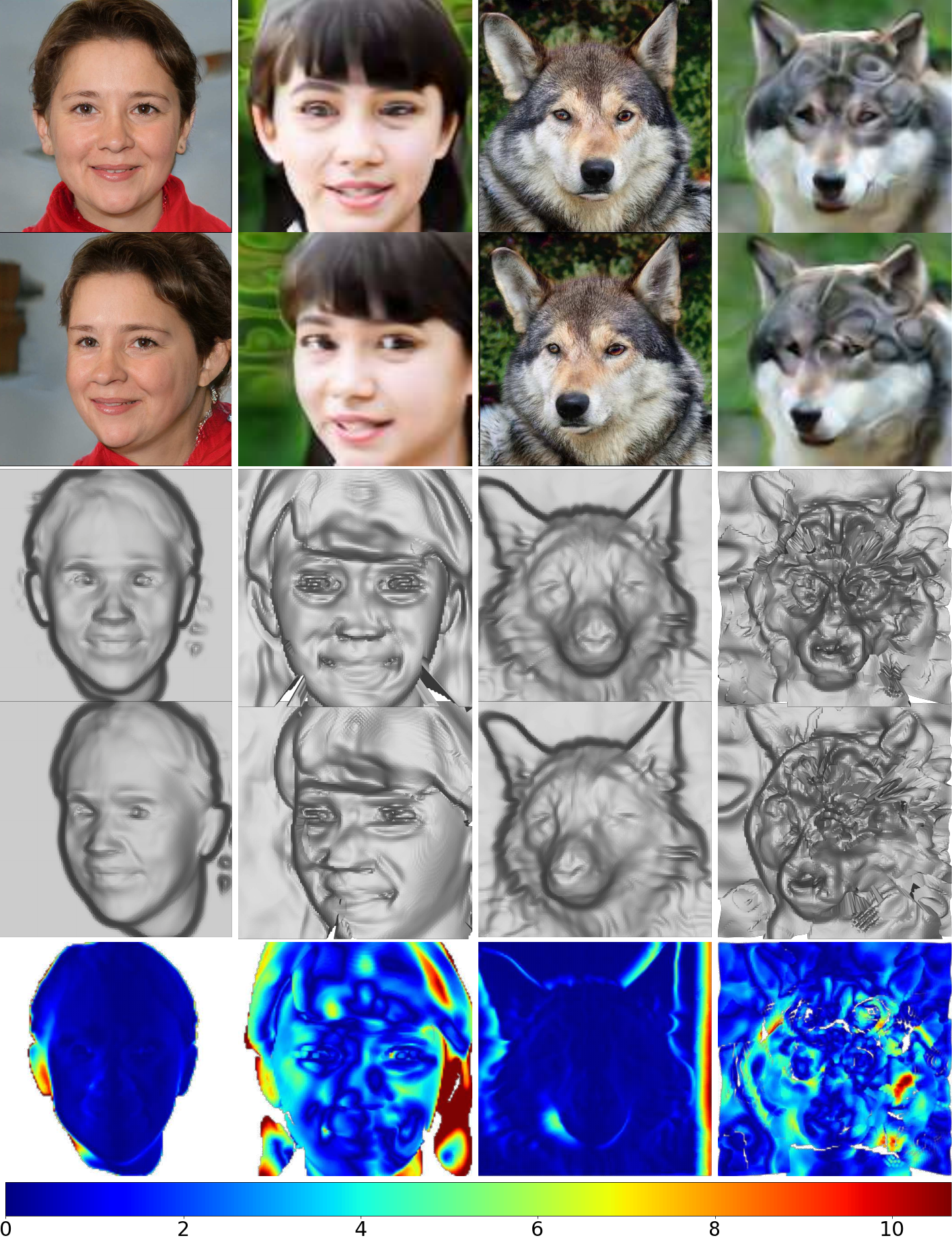}}
    \end{tabular}
    \vspace{-1em}
    \caption{Visual comparison of depth consistency. We visualize the nearest neighbour distances (in sample bin units) from the frontal depth maps to side-view depth maps. Our SDF-based technique significantly improves depth consistency compared to the baseline.
    }
    \label{fig:depth_consistency_viz}
    \vspace{-0.5cm}
\end{figure}

As shown in~\Cref{tbl:depth_consistency}, our use of SDF representation  dramatically improves depth consistency compared to the strongest current baseline PiGAN~\cite{chan2021pi}. ~\Cref{fig:depth_consistency_viz} shows the sample depth map pairs used for the evaluation and the error visualizations (in terms of distance to the closest point). %
The color map shows that our depth maps align well except for the occluded regions and backgrounds. In contrast, PiGAN depth maps show significant noise and spurious concave regions (e.g., nose of the dog).

\begin{figure} 
\includegraphics[width=1.0\linewidth]{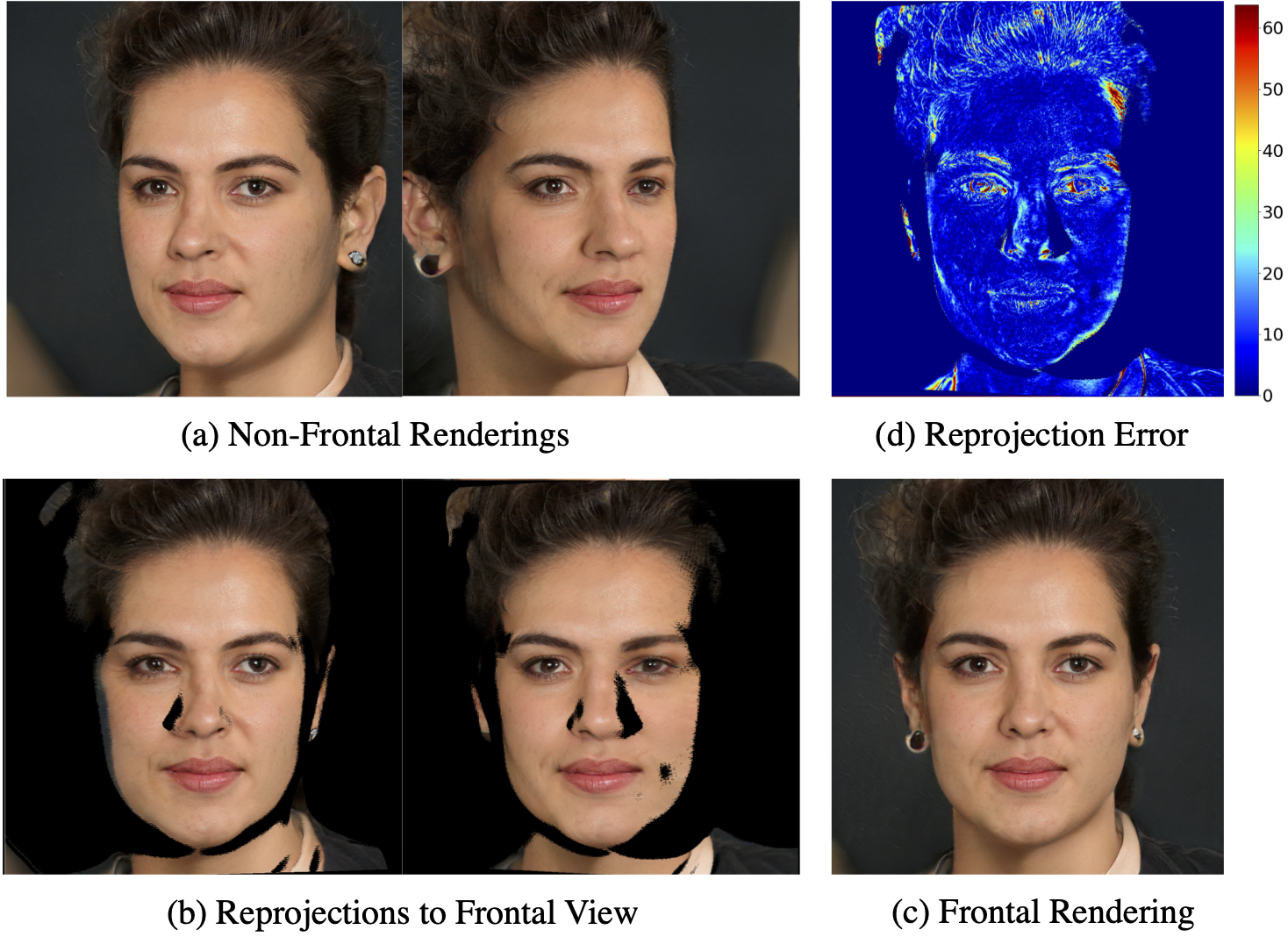}
\vspace{-0.6cm}
\caption{RGB rendering view-consistency. We render two side views (a) and project them to the frontal view (b) using the depth maps rendered from each views, ignoring occluded pixels. Note the high simmilarity between the reprojected images and the rendered frontal view (c), as can be seen from the error map (d). The error map shows mean absolute pixel difference for RGB channels (0-255) for the right side-view image. The errors are mostly from regions with high frequency textures and geometry (e.g., ear, hair), or  occlusion boundaries (right forehead).}\label{fig:reprojection}
\end{figure}

Moreover, we show that our consistent volume rendering naturally leads to high view-consistency for our RGB renderings. As shown in Fig.~\ref{fig:reprojection}, we visualize the reprojection of side-view renderings to the frontal view, using the depth values from volume rendering. The reprojected pixels closely match those of the original frontal view, indicating that our high-res multi-view RGB renderings and depth maps are all consistent to each other. Refer to supplementary for more detailed experiments.

\section{Limitations \& Future Work}
\label{sec:lim}

\begin{figure}[t!]
    \centering
    \includegraphics[width=1\linewidth]{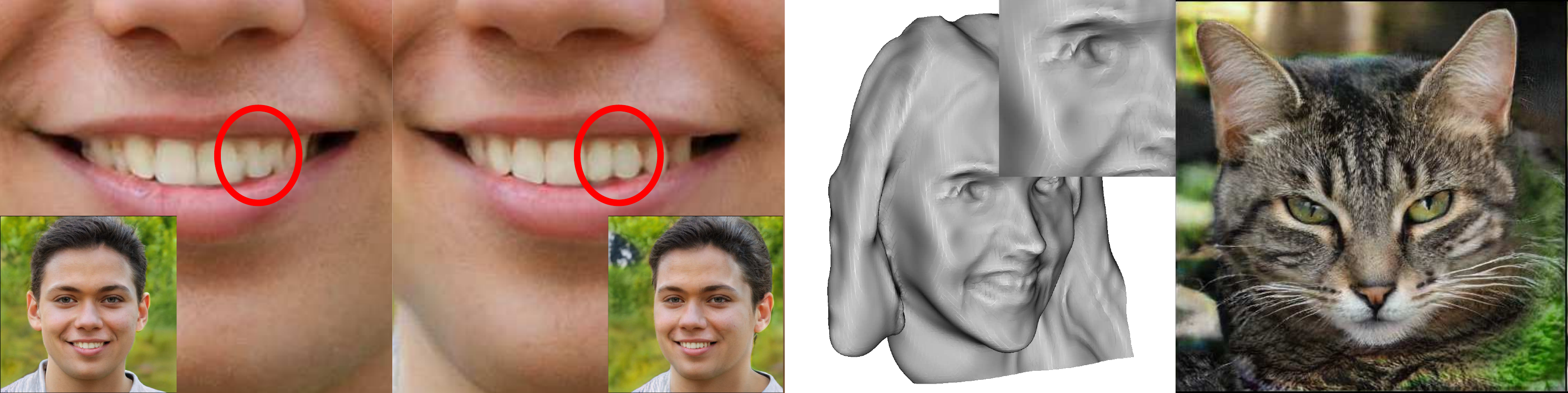}
    \caption{Limitations: potential aliasing artifacts, e.g., in teeth (left two columns). Specularities and shadows may create artifacts (3rd column from the left, cheek and eyes area), high curvatures are enhanced with radiance scaling filter~\cite{Vergne2010RadianceSF}. Inconsistencies in background might decrease photorealism (right column).}
    \label{fig:limitations}
\end{figure}

StyleSDF might exhibit minor aliasing and flickering, e.g., in teeth area. We leave it for future work since we expect those two to be corrected similarly to Mip-NeRF~\cite{Barron_2021_ICCV} and Alias-free StyleGAN~\cite{karras2021alias}. See example at left two columns of~\Cref{fig:limitations}. %
Specularities or other strong lighting effects currently introduce depth dents since StyleSDF might find it hard to disambiguate with no multi-view data (\Cref{fig:limitations} third column from the left). Adjusting the losses to include those effects is left for future work. Similarly, we do not currently separate foreground from background and use a single SDF for the whole image. \Cref{fig:limitations} (right column) shows how the cat's face is rendered properly, but the transition to the background is too abrupt, potentially diminishing photorealism. A potential solution could be adding an additional volume renderer to model the background as suggested in NeRF++~\cite{zhang2020nerf++}.

Finally, one may consider two improvements to the algorithm.  First one is training the two parts as a single end-to-end framework, instead of the current two networks.  In such case the StyleGAN2 discriminator would send proper gradients back to the volume renderer to produce optimal feature maps, which might lead to even more refined geometry. However, end-to-end training poses a trade-off. The increased GPU memory consumption of this setup would require either a decreased batch size, which might hurt the overall performance, or increased training time if we keep the batch size and accumulate gradients. Second improvement could be to create a volume sampling strategy 
tied to SDF's surface boundary (to reduce the number of query points at each forward pass) and eliminate the need for a 2D CNN that upsamples feature maps. That would tie 3D geometry directly to the high resolution image. 

\section{Conclusions}
\label{sec:conc}
We introduced StyleSDF, a method that can render 1024x1024 view-consistent images along with the detailed underlying  geometry. 
The proposed architecture combines SDF-based volume renderer and a 2D StyleGAN network and is trained to generate realistic images for all sampled viewpoints via adversarial loss, naturally inducing view-consistent 3D scenes. StyleSDF  represents and learns  complex 3D shape and appearance without  multi-view or 3D supervision, requiring only a dataset of single-view images, suggesting a new route ahead for neural 3D content generation, editing, and reconstruction.

\section*{Acknowledgements}
We wish to thank Aleksander Holynski for his valuable advice. 
This work was supported by the UW Reality Lab, Meta, Google, Futurewei, Amazon and Adobe. J.J. Park was supported by the Apple fellowship.

{\small
\bibliographystyle{ieee_fullname}
\bibliography{egbib}
}

\ifarxiv
\clearpage
\appendix
\section*{Appendix}
In this appendix we provide additional qualitative and quantitative results on our approach, along with the technical details that supplement the main paper. In Sec.~\Cref{sec: societal}, we discuss possible societal impacts of our technology. Then, we present additional experiments on the view-consistency of our RGB renderings via image reprojection (Sec.~\Cref{Sec: view_consist}). We further demonstrate the quality of our 3D shapes in Sec.~\Cref{sec: qual_3d}. In Sec.~\Cref{sec: video}, we describe the content of the supplementary videos and introduce a geometry-aware noise injection procedure to reduce flickering. Next, we discuss implementation details and the merits of our proposed sampling strategy in (Sec.~\Cref{sec: implementation} and~\Cref{sec:sampling_strategy} respectively. We conduct ablation studies on our approach in ~\Cref{sec:ablations}). Finally, we continue our discussion on our method's limitations in Sec.~\Cref{sec: limitation}.

\section{Societal Impacts}\label{sec: societal}
Image and 3D model-generating technologies (e.g., deepfakes) could be used for spreading misinformation about existing or non-existing people \cite{hill2020designed, kim2018deep}. Our proposed technology allows generating multi-view renderings of a person, and might be used for creating more realistic fake videos. These problems could potentially be addressed by developing algorithms to detect neural network-generated content \cite{wang2020cnn}. We refer readers to \cite{tewari2020state} for strategies of mitigating negative social impacts of neural rendering. Moreover, image generative models are optimized to follow the training distribution, and thus could inherit the ethnic, gender, or other biases present in the training dataset. A possible solution is creating a more balanced dataset, e.g., as in \cite{karkkainen2021fairface}.

\section{View Consistency of RGB Renderings}\label{Sec: view_consist}
\subsection{Volume Rendering Consistency}
The consistent volume rendering from our SDF-based technique naturally leads to high view consistency of our RGB renderings. 
To show the superior 3D-consistency of our SDF-based volume rendering, we measure the reprojection error when a side view pixels are warped to the frontal view. 
We randomly sample 1,000 identities and render the depth and RGB images at $256\times 256$ and set the side view to be $1.5\times$ the standard deviation of the azimuth distribution in training (which is 0.45 radians for FFHQ and 0.225 radians for AFHQ). 
We reproject the side-view RGB renderings to the frontal view using the side-view depth, and we do not ignore occluded pixels. We measure color inconsistency with the median of pixel-wise L1 error in RGB (0 - 255), averaged over the 1,000 samples. The use of median effectively removes the large errors coming from occlusions. Note that since PiGAN is trained with center-cropped FFHQ images (resized to $320\times 320$ and center-cropped to $256\times 256$), we apply the same transformation on our results before computing the median.

\begin{table}[t]
\centering
    \begin{tabular}{lrr}
    \toprule
    Dataset:     & FFHQ & AFHQ \\
    \midrule
    PiGAN~\cite{chan2021pi}        & 14.7    & 16.5    \\
    Ours (volume renderer)         & \textbf{2.9}    & \textbf{2.6}    \\
    \bottomrule
    \end{tabular}

\caption{Quantitative view-consistency comparison of the RGB renderings. We evaluate the color error of the RGB volume renderings between the frontal view and the reprojection from a fixed side view. The error is measured as the median of the per-pixel mean absolute difference (0 - 255). We average the color inconsistency over 1,000 samples for each dataset. Our underlying SDF geometry representation promotes superior 3D consistency. (also see \cref{fig:color_consistency}).}
\label{tbl:color_consistency}
\ifarxiv
\vspace{-0.25cm}
\fi
\end{table}

\begin{figure*}[t]
    \centering
    \setlength{\tabcolsep}{0pt}
    \includegraphics[width=\linewidth]{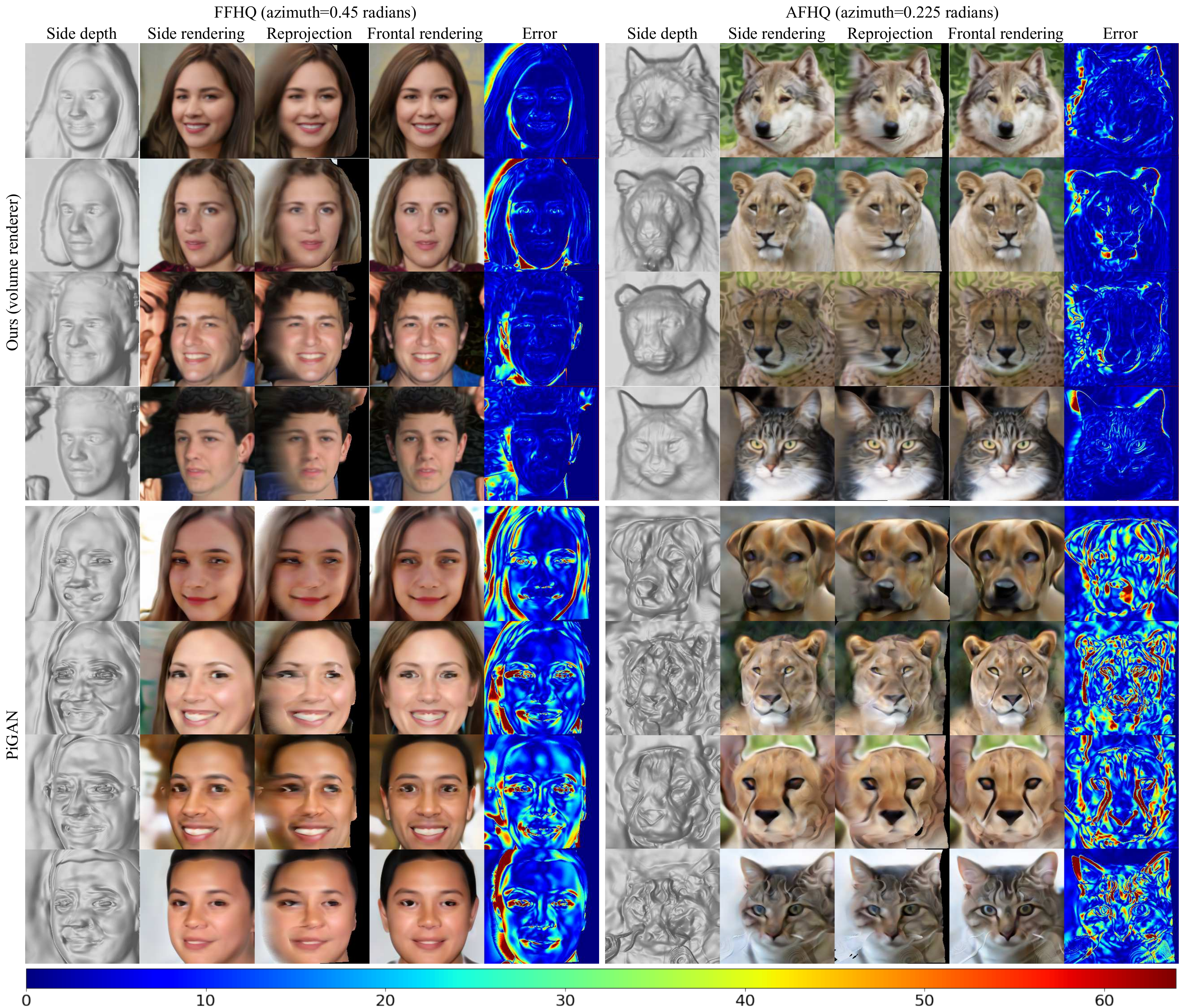}
    \caption{
    Qualitative view consistency comparison of RGB renderings. 
    We project the rendering from a side view using its corresponding depth map to the frontal view. We compare the reprojection to the frontal-view rendering and compute the error map showing mean absolute difference in RGB channels (0 - 255).
    Our SDF-based technique generate superior depth quality and significantly improves the view-consistency of the RGB renderings. Most of our errors concentrate on the occlusion boundaries whereas PiGAN's errors spread across the whole subject (e.g., eyes, mouth, specular highlights, fur patterns).}
    \label{fig:color_consistency}
    \ifarxiv
    \vspace{-0.4cm}
    \fi
\end{figure*}

As shown in \cref{tbl:color_consistency}, 
StyleSDF presents significantly improved color consistency
 compared to the strongest current baseline, PiGAN~\cite{chan2021pi}. \cref{fig:color_consistency} shows the sample depth and color rendering pairs used for the evaluation, along with the pixelwise error maps. The error maps demonstrate that our volume RGB renderings have high view consistency, as the large reprojection errors are mostly in the occluded regions. On the other hand, PiGAN's reprojections do not align well with the frontal view, showing big errors also near the eyes, mouth, in presence of specular highlights, etc.

\begin{figure*}[t!]
    \centering
    \setlength{\tabcolsep}{0pt}
    \def\imW{0.0769\textwidth}
    \begin{tabular}{>{\centering\arraybackslash}m{\imW}
                    >{\centering\arraybackslash}m{\imW}
                    >{\centering\arraybackslash}m{\imW}
                    >{\centering\arraybackslash}m{\imW}
                    >{\centering\arraybackslash}m{\imW}
                    >{\centering\arraybackslash}m{\imW}
                    >{\centering\arraybackslash}m{\imW}
                    >{\centering\arraybackslash}m{\imW}
                    >{\centering\arraybackslash}m{\imW}
                    >{\centering\arraybackslash}m{\imW}
                    >{\centering\arraybackslash}m{\imW}
                    >{\centering\arraybackslash}m{\imW}
                    >{\centering\arraybackslash}m{\imW}}
          \scalebox{0.7}{Frontal} &   \scalebox{0.6}{$\phi = -0.45$}  & \scalebox{0.6}{$\phi = -0.3$} &  \scalebox{0.6}{$\phi = -0.15$} & \scalebox{0.6}{$\phi = 0.15$} & \scalebox{0.6}{$\phi = 0.3$}  & \scalebox{0.6}{$\phi = 0.45$} & \scalebox{0.6}{$\theta = -0.225$} & \scalebox{0.6}{$\theta = -0.15$} & \scalebox{0.6}{$\theta =-0.075$} & \scalebox{0.6}{$\theta = 0.075$} & \scalebox{0.6}{$\theta = 0.15$} & \scalebox{0.6}{$\theta = 0.225$}\tabularnewline
        \multicolumn{13}{p{\textwidth}}{\includegraphics[width=\textwidth]{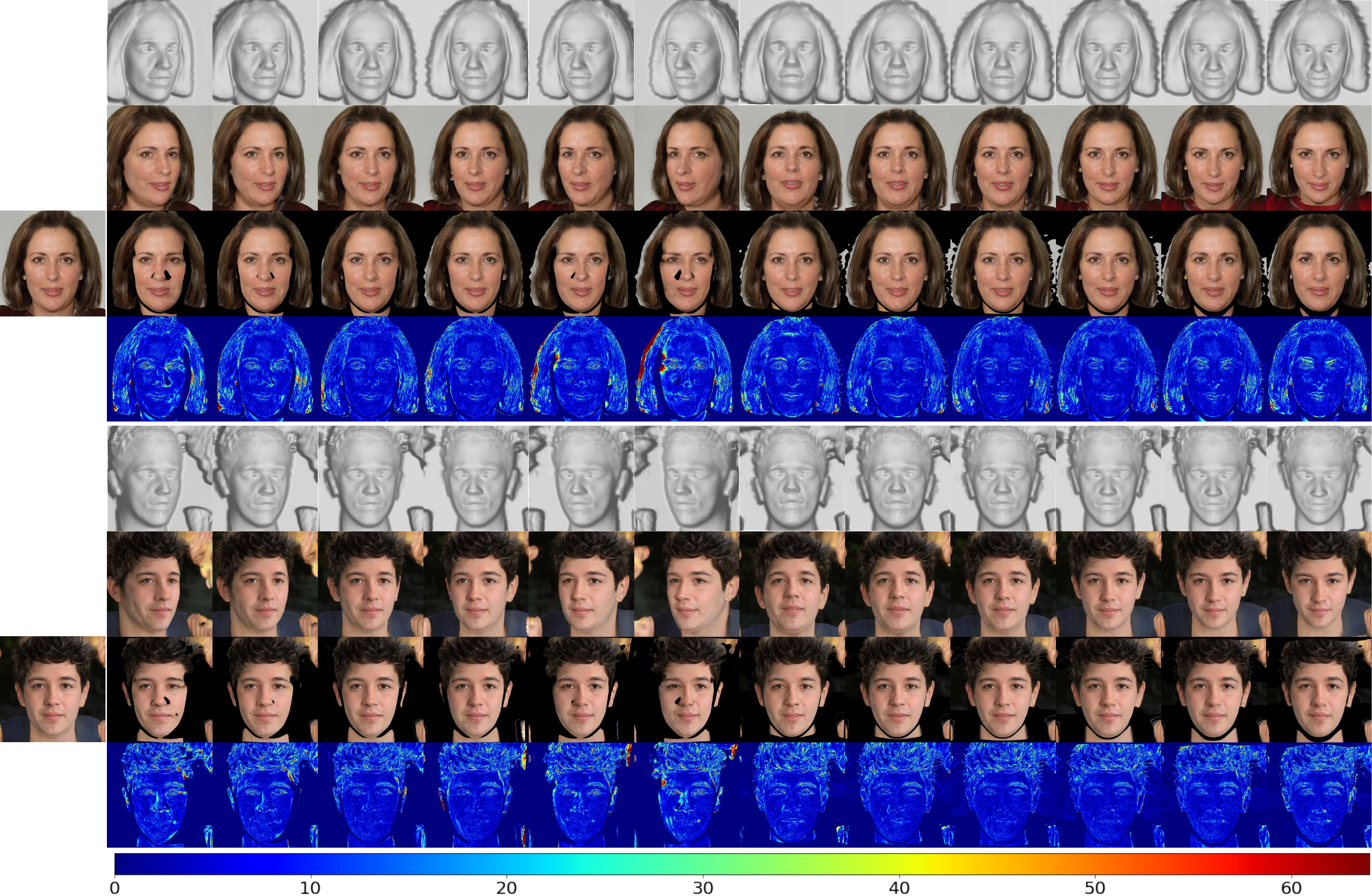}}
    \end{tabular}
    \caption{View-consistency visualization of high-resolution renderings. We use the side-view depth maps (first rows) to warp the side-view RGB renderings (second rows) to the frontal view (first column). The reprojected pixels that pass the occlusion testing are shown in the third row. We compare the reprojections with the frontal-view renderings and show the per-pixel error maps (fourth rows). Our reprojections well align with the frontal view with errors mostly in the occlusion boundaries and high-frequency details.}
    \label{fig:color_consistency_multiview}
    \ifarxiv
    \vspace{-0.4cm}
    \fi
\end{figure*}

\subsection{High-Resolution RGB Consistency}
In \cref{fig:color_consistency_multiview}, we present the reprojection experiment results using our high-resolution RGB outputs. As in the volume rendering consistency experiment, we reproject the RGB pixels from non-frontal views (with varying azimuth and elevation) to the frontal views. The results demonstrate the strong 3D-consistency of our high-resolution images, as the reprojected non-frontal images are similar to the frontal renderings. However, as mentioned in the limitation section of the main paper, the current implementation of StyleGAN2 comes with significant aliasing of the high-frequency components, resulting in noticeable pixel errors on regions with high-frequency details, e.g., hair, ears, eyes, etc. 
To identify the errors in the high-frequency details, we visualize the mean reprojection images. I.e. we project non-frontal views and average the pixel values across views. As can be seen in \cref{fig:color_consistency_mean}, the mean reprojection images closely replicate the identities and important structures of the frontal view, demonstrating strong view-consistencies. The error map confirms that most of the errors are concentrated on the high-frequency noise of the StyleGAN generator.

\begin{figure*}[t]
    \centering
    \includegraphics[width=0.82\linewidth]{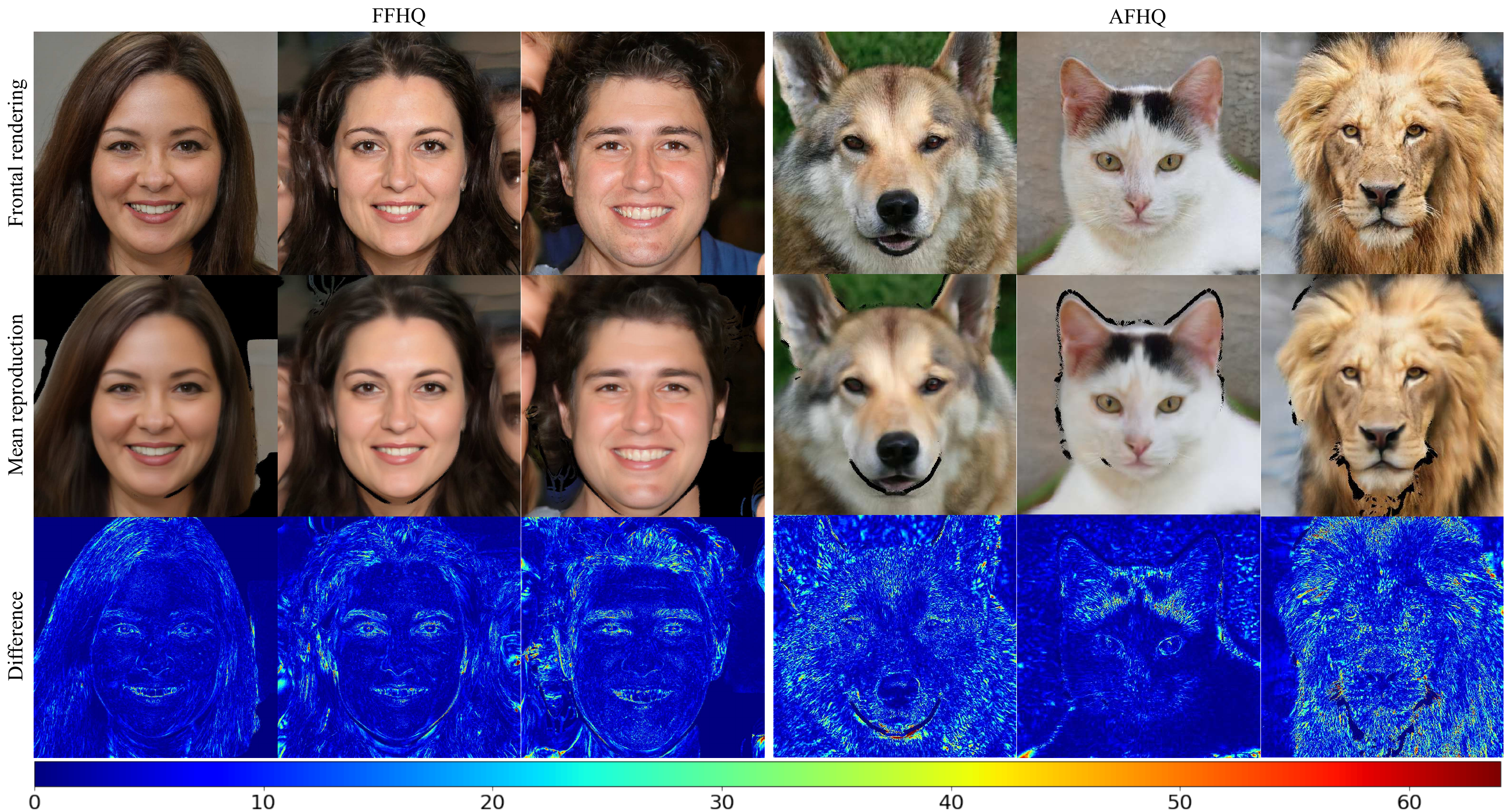}
    \caption{Color consistency visualization with mean faces. We reproject the RGB renderings from the side views to the frontal view  (as in \cref{fig:color_consistency_multiview}). We show the mean reprojections that pass the occlusion testing and their differences to the frontal-view renderings. The mean reprojections are well aligned with the frontal rendering. The majority of the errors are in the high-frequency details, generated from the random noise maps in the StyleGAN component. This demonstrates the strong view consistency of our high-resolution renderings.}
    \label{fig:color_consistency_mean}
\end{figure*}

\section{Qualitative 3D results}\label{sec: qual_3d}
We demonstrate the consistency of our 3D representation by overlaying the point clouds from the frontal and side view depth maps (\cref{subfig:extracted_meshes_overlay}). The visualization, shown in two different colors, clearly shows high consistency between the depth maps. To show the quality and plausibility of our 3D models, we extract meshes on our SDFs via marching cubes and visualize them in extreme angles (\cref{subfig:extracted_meshes_marching_cubes}).

\begin{figure*}[t!]
\centering
    \setlength\tabcolsep{0pt}
    \begin{tabular}{>{\centering\arraybackslash}m{0.2\linewidth}
                    >{\centering\arraybackslash}m{0.2\linewidth}
                    >{\centering\arraybackslash}m{0.2\linewidth}
                    >{\centering\arraybackslash}m{0.2\linewidth}
                    >{\centering\arraybackslash}m{0.2\linewidth}}
    \includegraphics[width=\linewidth]{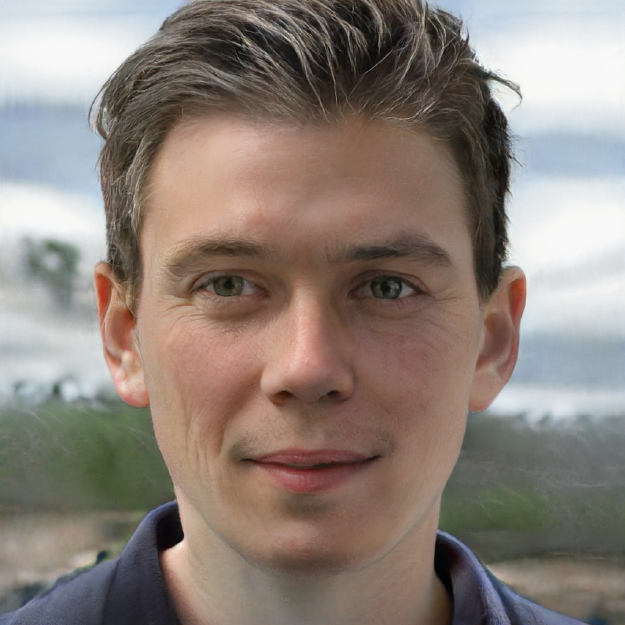} &
    \includegraphics[width=\linewidth]{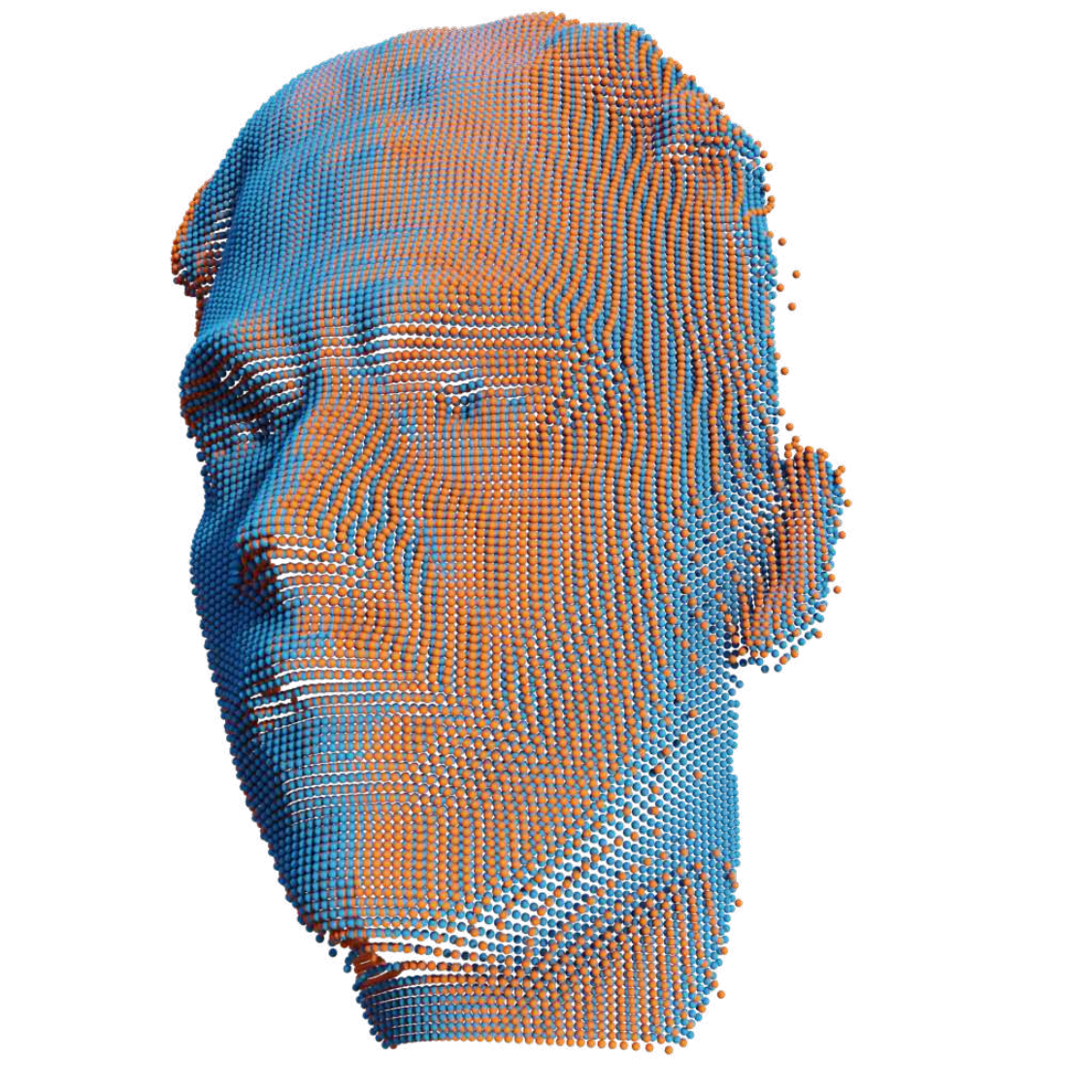} &
    \includegraphics[width=\linewidth]{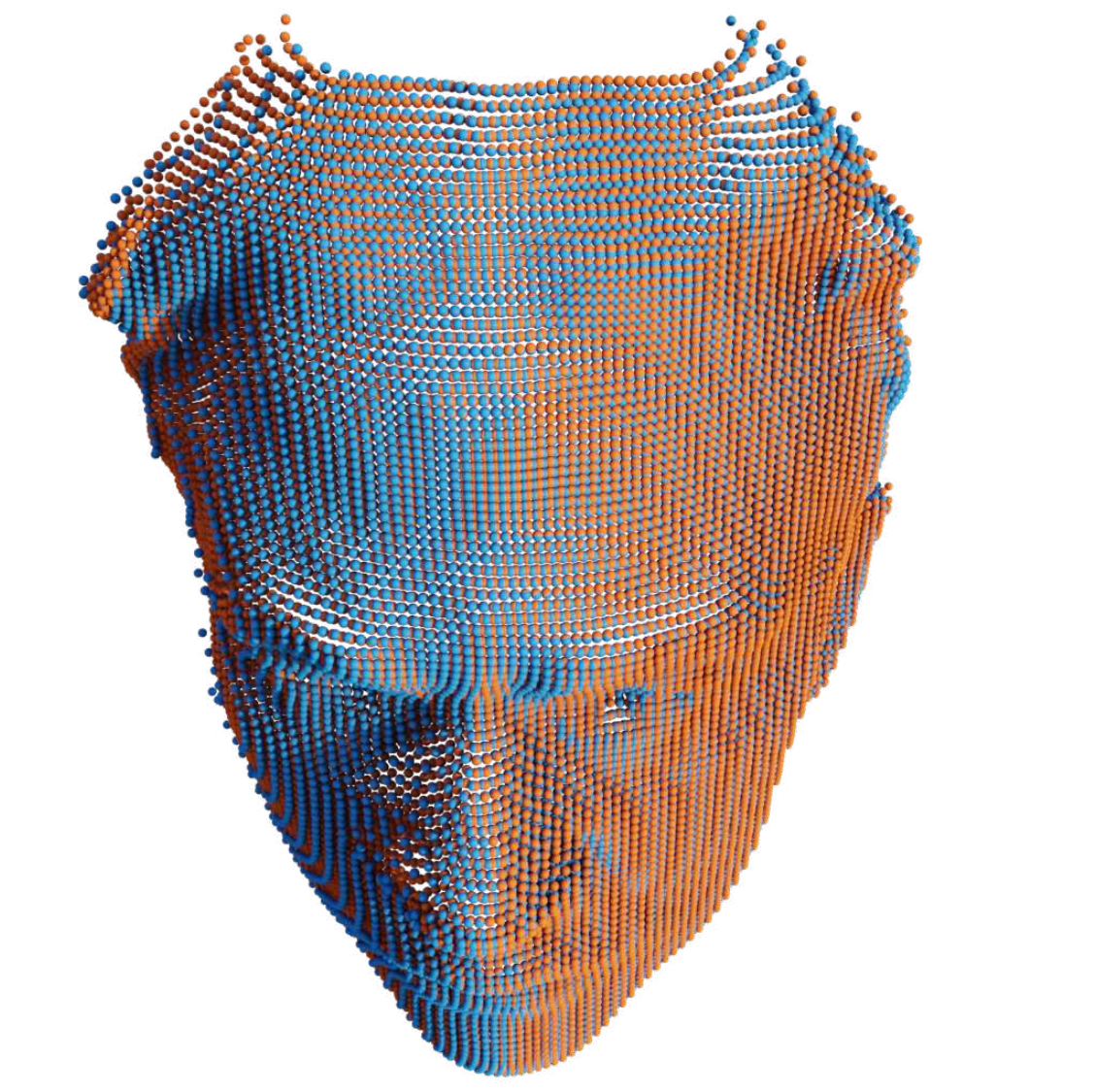} &
    \includegraphics[width=\linewidth]{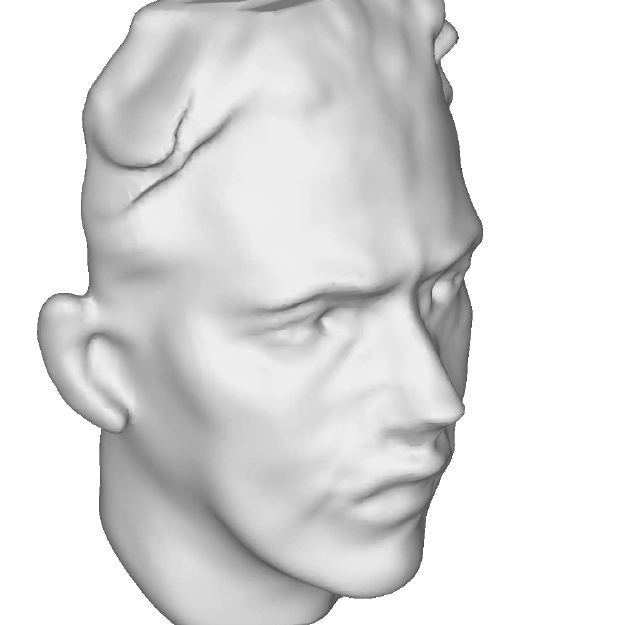} &
    \includegraphics[width=\linewidth]{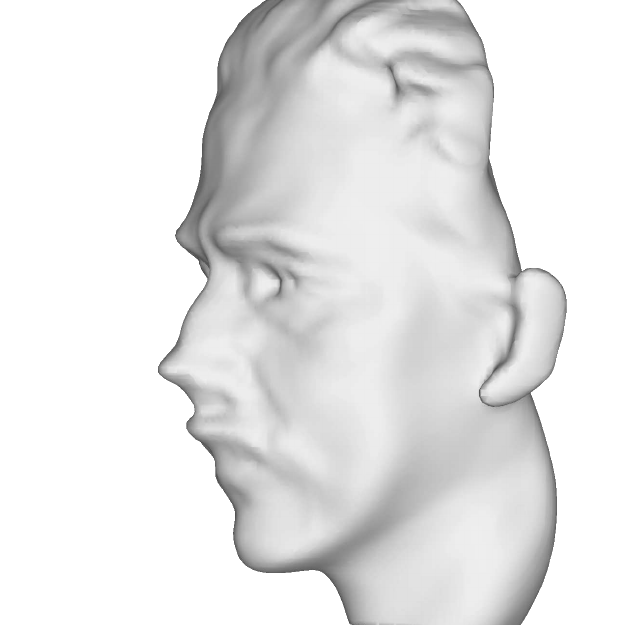}\tabularnewline%
    \includegraphics[width=\linewidth]{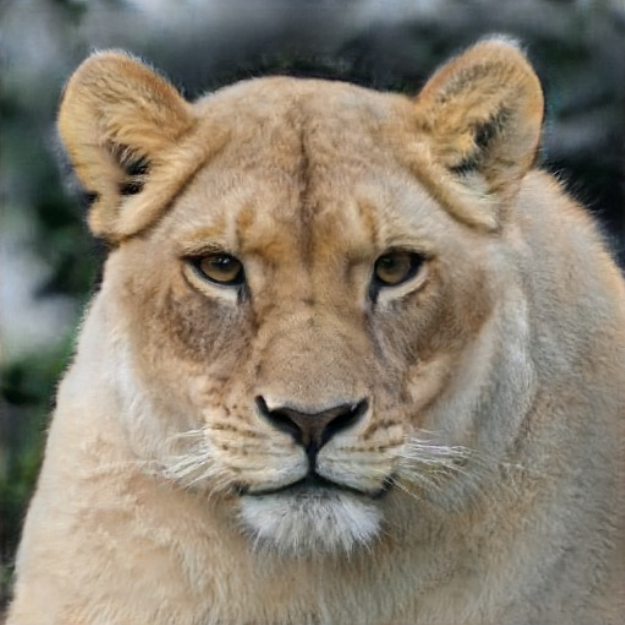} &
    \includegraphics[width=\linewidth]{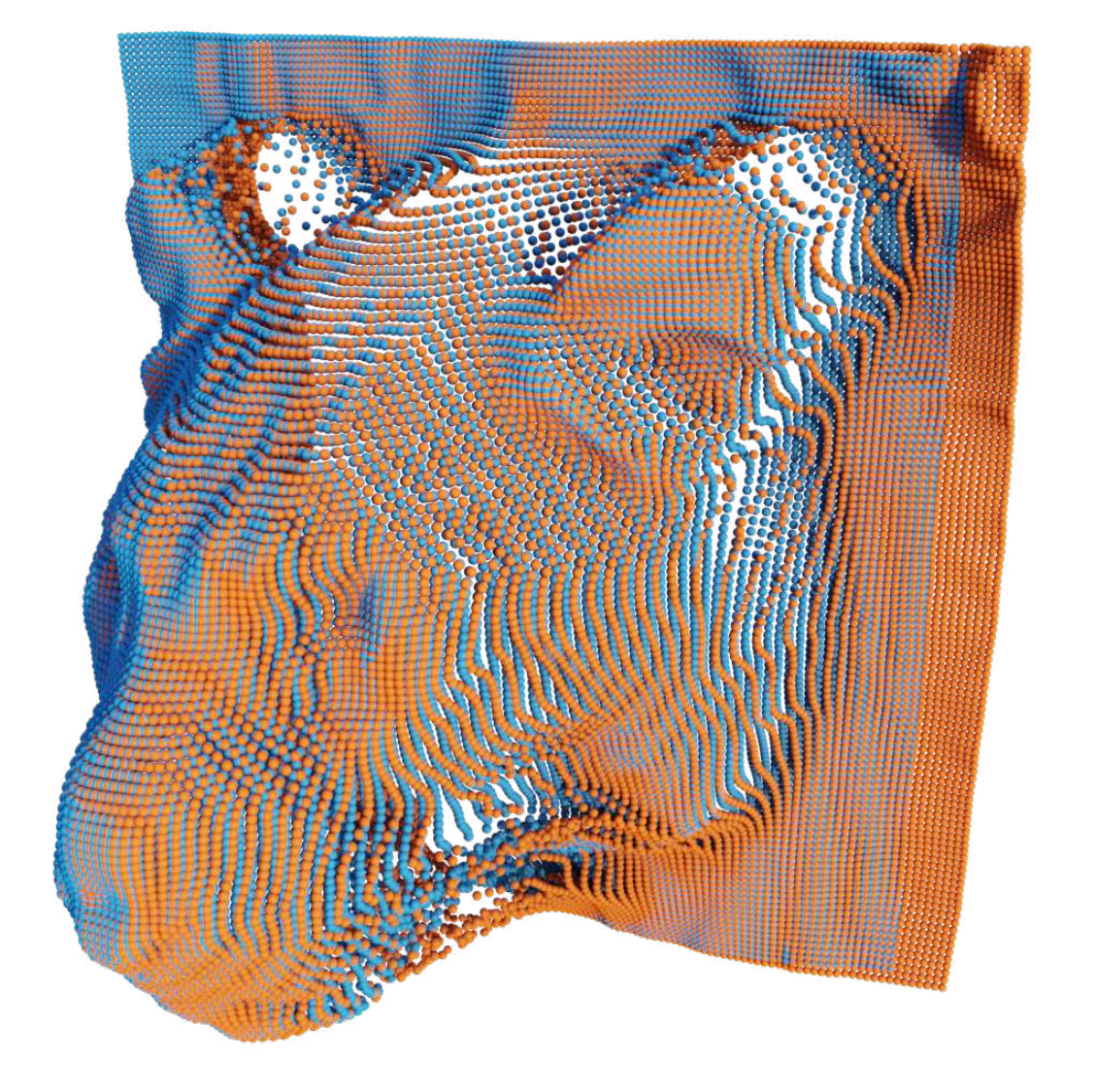} &
    \includegraphics[width=\linewidth]{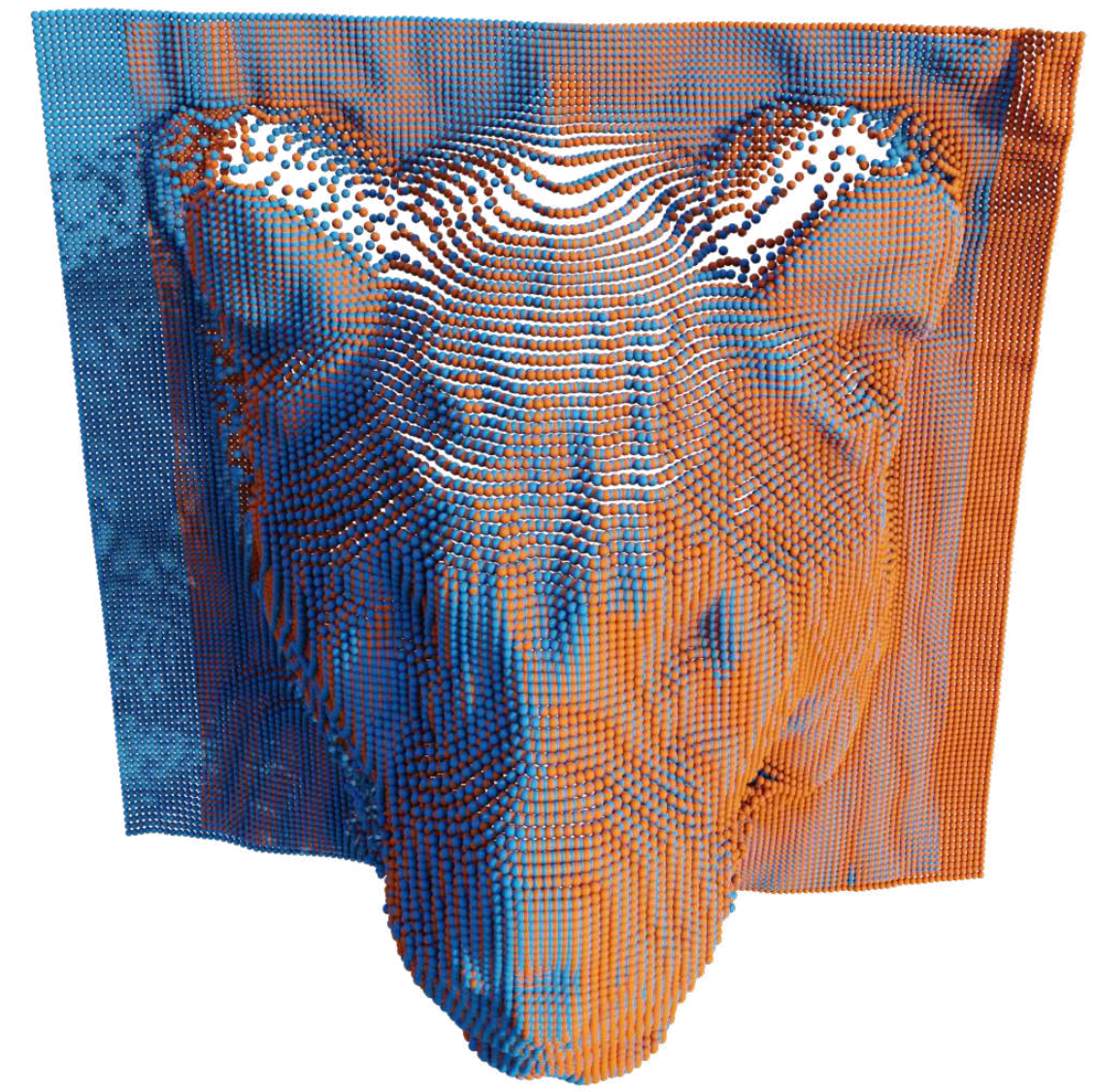} &
    \includegraphics[width=\linewidth]{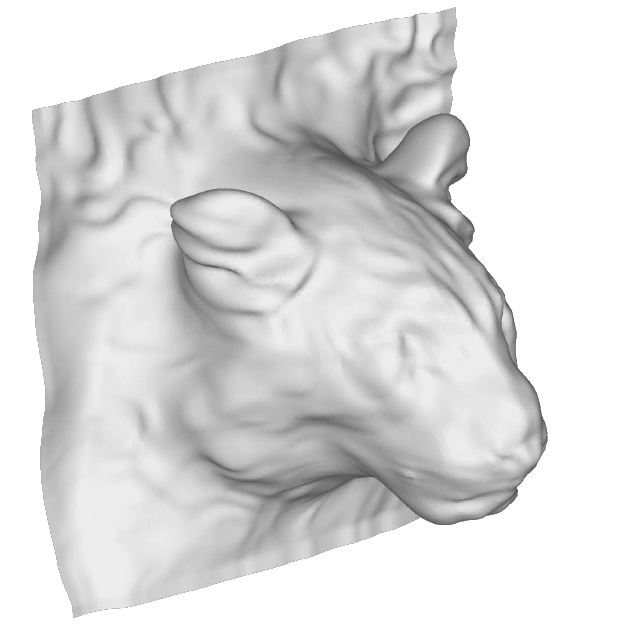} &
    \includegraphics[width=\linewidth]{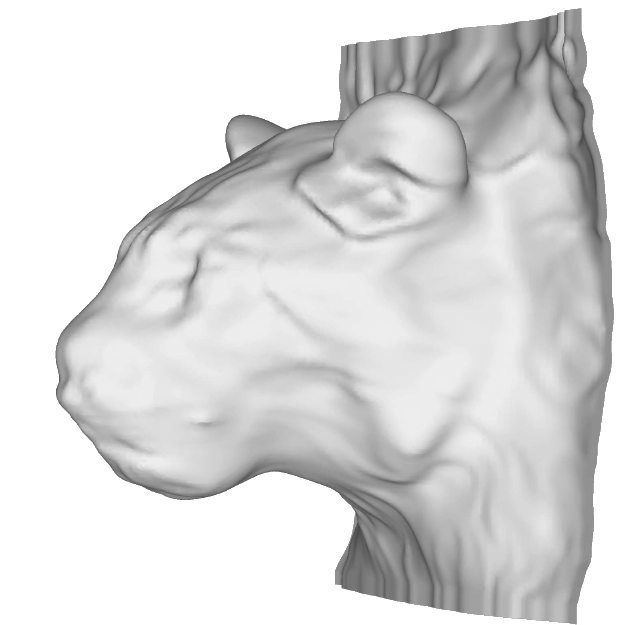}\tabularnewline%
    \begin{subfigure}{\linewidth}
        \centering 
        \caption{\scriptsize{Color}}
    \end{subfigure} & 
    \multicolumn{2}{m{0.4\linewidth}}{
        \begin{subfigure}{\linewidth}
            \centering 
            \caption{\scriptsize{Overlaid Depth Maps}}
            \label{subfig:extracted_meshes_overlay}
        \end{subfigure}} & 
    \multicolumn{2}{m{0.4\linewidth}}{
        \begin{subfigure}{\linewidth}
            \centering 
            \caption{\scriptsize{Extracted Marching Cubes}}
            \label{subfig:extracted_meshes_marching_cubes}
        \end{subfigure}}
    \end{tabular}
\vspace{-0.4cm}
\caption{Consistent and plausible 3D shapes. (a) Color images. (b) Overlaid point clouds extracted from frontal and side view depth maps. (c) Marching cubes meshes, rendered from extreme angles.}
\label{fig:extracted_meshes}
\vspace{-0.5cm}
\end{figure*}

\section{Video Results}\label{sec: video}
Since our 3D-consistent high-resolution image generation can be better appreciated with videos, we have attached 24 sequences in the supplementary material, featuring view-generation results on the two datasets using two different camera trajectories. For each identity, we provide two videos, one for RGB and another for depth rendering. The videos are presented in the \href{https://stylesdf.github.io/}{project's website}.

\subsection{Geometry-Aware StyleGAN Noise}
Even though the images shown in the main paper on multi-view RGB generation look highly realistic, we note that for generating a video sequence, the random noise of StyleGAN2 \cite{karras2019style}, when na\"ively applied to 2D images, could result in severe flickering of high-frequency details between frames. The flickering artifacts are especially prominent for the AFHQ dataset due to high-frequency textures from the fur patterns.

Therefore, we aim at reducing this flickering by adding the Gaussian noise in a 3D-consistent manner, i.e., we want to attach the noise on the 3D surface. We achieve this by extracting a mesh (at 128 resolution grid) for each sequence from our SDF representation and attach a unit Gaussian noise to each vertex, and render the mesh using vertex coloring. Since higher resolution intermediate features require up to 1024$\times$1024 noise map, we subdivide triangle faces of the extracted mesh once every layer, starting from 128$\times$128 feature layers. 
The video results show that the geometry-aware noise injection reduce the flickering problem on the AFHQ dataset, but noticeable flickering still exist. Furthermore, we observe that the geometry-aware noise slightly sacrifices individual frame's image quality, presenting less pronounced high-frequency details, likely due to the change of the Gaussian noise distribution during the rendering process. The videos rendered with geometry-aware noise can be viewed at the \href{https://stylesdf.github.io/}{project's website}.

\begin{figure*}[t!]
\centering
    \setlength\tabcolsep{0pt}
    \begin{tabular}{>{\centering\arraybackslash}m{0.25\linewidth}
                    >{\centering\arraybackslash}m{0.25\linewidth}
                    >{\centering\arraybackslash}m{0.25\linewidth}
                    >{\centering\arraybackslash}m{0.25\linewidth}}
    \includegraphics[width=\linewidth]{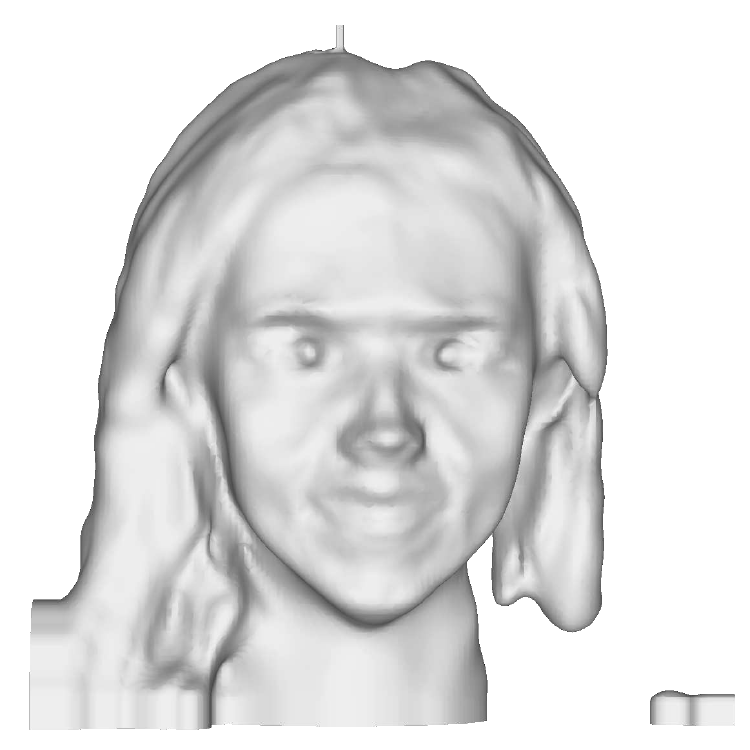} &
    \includegraphics[width=\linewidth]{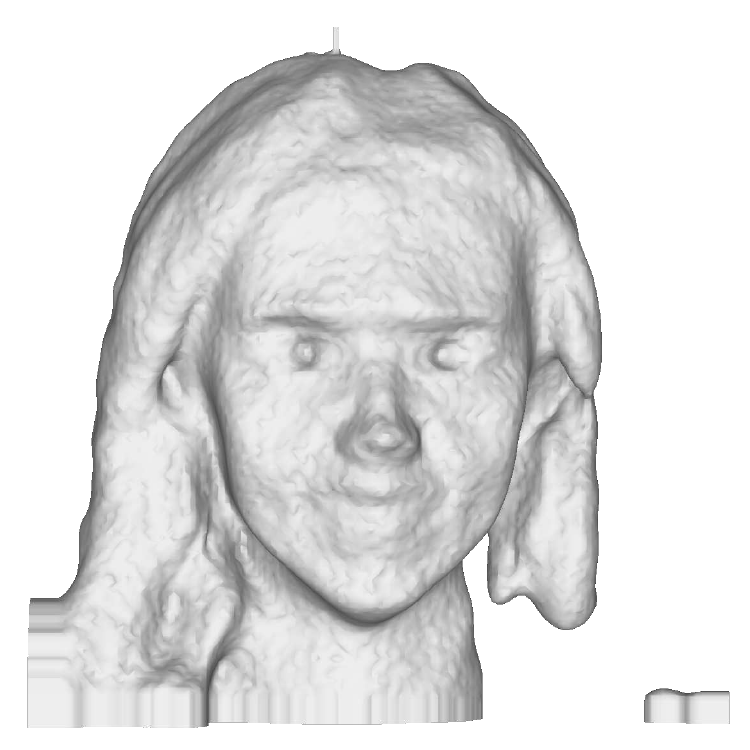} &
    \includegraphics[width=\linewidth]{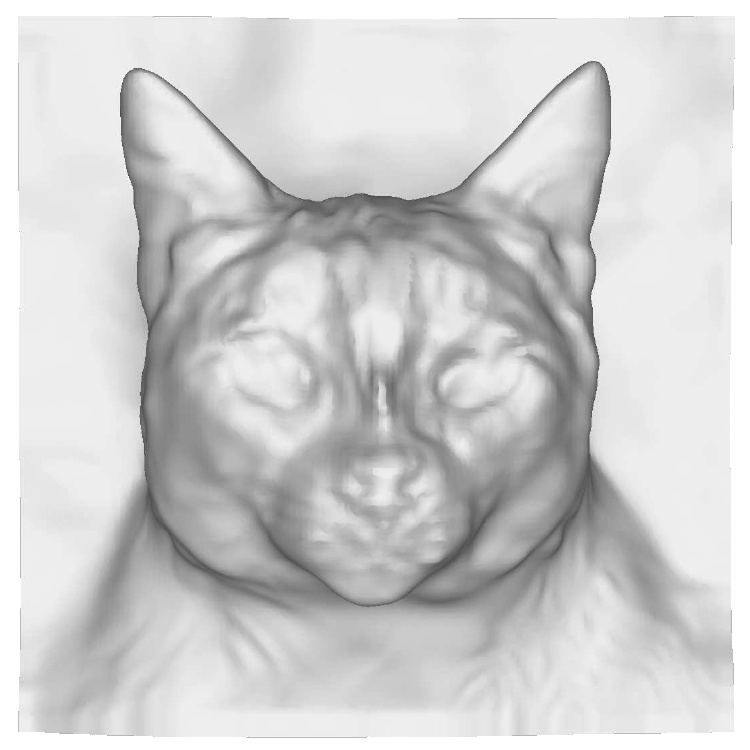} & 
    \includegraphics[width=\linewidth]{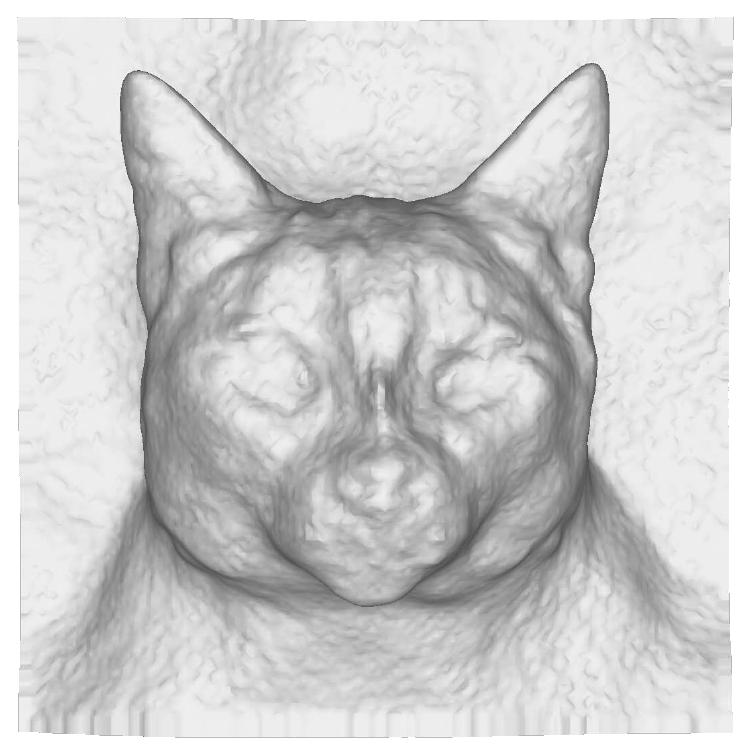} \tabularnewline[-0.3cm]
    \scriptsize{Ours} & \scriptsize{Stratified} & \scriptsize{Ours} & \scriptsize{Stratified}
    \end{tabular}
\caption{We compare extracted meshes using our sampling strategy vs. stratified sampling. Note the noise induced by stratified sampling. (zoom in for details)}
\label{fig:stratified_ablation}
\end{figure*}

\section{Implementation Details} \label{sec: implementation}
\subsection{Dataset Details}
\vspace{0.1cm}\noindent\textbf{FFHQ:} We trained FFHQ with R1 regularization loss of $10$. The camera field of view was fixed to $12^{\circ}$ and its azimuth and elevation angles are sampled from Normal distributions with zero mean and standard deviations of $0.3$ and $0.15$ respectively. We set the near and far fields to $[0.88,1.12]$ and sample 24 points per ray during training .We trained our volume renderer for $200k$ iterations and the 2D-Styled generator for $300k$ iterations. 

\vspace{0.1cm}\noindent\textbf{AFHQ:} The AFHQ dataset contains training and validation sets for 3 classes, cats, dogs and wild animals. We merged all the training data into a single training set. We apply R1 regularization loss of $50$. Both azimuth and elevation angles are sampled from a Gaussian distribution with zero mean and standard deviation of $0.15$ and a camera field of view of $12^{\circ}$. The near and far fields as well as the number of samples per ray are identical to the FFHQ setup. Our volume renderer as well as the 2D-Styled generator were trained for $200k$ iterations.

\begin{figure*}[th!]
\centering
    \setlength\tabcolsep{0pt}
    \begin{tabular}{>{\centering\arraybackslash}m{0.1\linewidth}
                    >{\centering\arraybackslash}m{0.15\linewidth}
                    >{\centering\arraybackslash}m{0.15\linewidth}
                    >{\centering\arraybackslash}m{0.15\linewidth}
                    >{\centering\arraybackslash}m{0.15\linewidth}
                    >{\centering\arraybackslash}m{0.15\linewidth}
                    >{\centering\arraybackslash}m{0.15\linewidth}}
    \scriptsize{Without Minimal Surface Loss} &
    \multicolumn{6}{m{0.9\linewidth}}{\includegraphics[width=\linewidth]{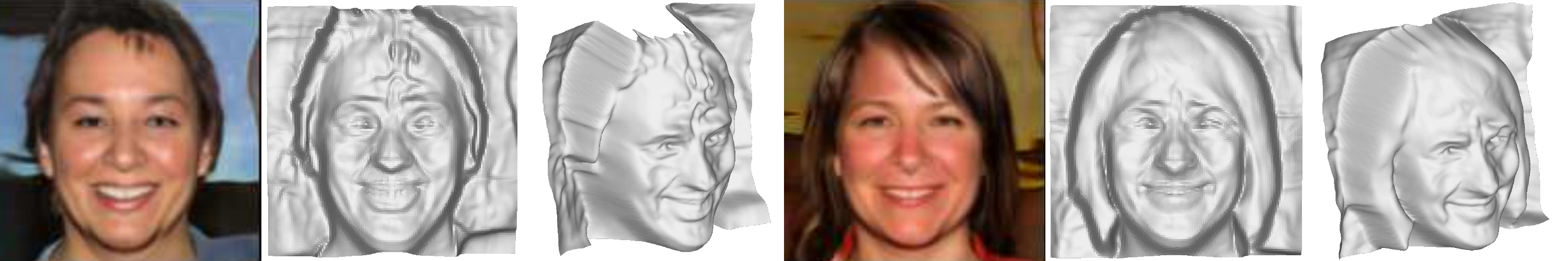}}\tabularnewline
    \scriptsize{With Minimal Surface Loss} &
    \multicolumn{6}{m{0.9\linewidth}}{\includegraphics[width=\linewidth]{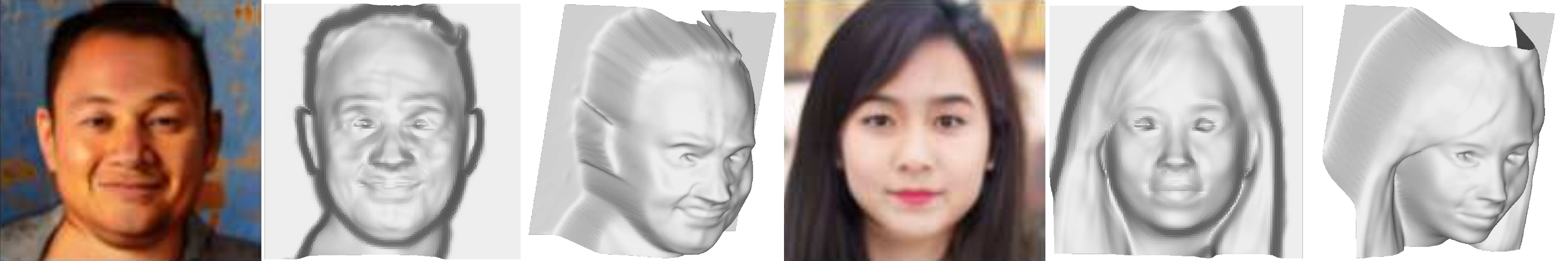}}\tabularnewline
    & \scriptsize{Color Image} & \scriptsize{Frontal view} & \scriptsize{Side view} & \scriptsize{Color Image} & \scriptsize{Frontal view} & \scriptsize{Side view}
    \end{tabular}
\caption{Minimal surface loss ablation study. 
We visualize the volume rendered RGB and depth images from volume renderers trained with and without the minimal surface loss. 
 The Depth map meshes are visualized from the front and side views. Note how a model trained with the minimal surface loss generates smoother surfaces and is less prone to shape-radiance ambiguities, e.g., specular highlights are baked into the geometry.}
\label{fig:min_surf_ablation}
\ifarxiv
\vspace{-0.4cm}
\fi
\end{figure*}

\subsection{Training Details}
\vspace{0.1cm}\noindent\textbf{Sphere Initialization:} During our experiments we have noticed that our SDF volume renderer can get stuck at a local minimum, which generates concave surfaces. To avoid this optimization failure, we first initialize the MLP to generate an SDF of a sphere centered at the origin with a fixed radius. We analytically compute the signed distance of the sampled points from the sphere and fit the MLP to match these distances. We run this procedure for $10k$ iterations before the main training. The importance of sphere initialization is discussed in~\Cref{sec:ablations}. 

\vspace{0.1cm}\noindent\textbf{Training setup:} Our system is trained in a two-stage strategy. First, we train the backbone SDF volume renderer on $64 \times 64$ images with a batch size of 24 using the ADAM~\cite{kingma2014adam} optimizer with learning rates of $2\cdot10^{-5}$ and $2\cdot10^{-4}$ for the generator and discriminator respectively and $\beta_1 =0, \beta2 = 0.9$. We accumulate gradients in order to fit to the GPU memory constraints. For instance, a setup of 2 NVIDIA A6000 GPUs (a batch of 12 images per GPU) requires the accumulation of two forward passes (6 images per forward pass) and takes roughly 3.5 days to train. We use an exponential moving average model during inference.

In the second phase, we freeze the volume renderer weights and train the 2D styled generator with identical setup to StyleGAN2~\cite{karras2020analyzing}. This includes ADAM optimizer with $0.002$ learning rate and $\beta_1 = 0, \beta_2 = 0.99$, equalized learning rate, lazy R1 and path regularization, batch size of 32, and exponential moving average. We trained the styled generator on 8 NVIDIA TeslaV100 GPUs for 7 days.

\begin{figure*}[th!]
\centering
    \setlength\tabcolsep{0pt}
    \begin{tabular}{>{\centering\arraybackslash}m{0.1\linewidth}
                    >{\centering\arraybackslash}m{0.15\linewidth}
                    >{\centering\arraybackslash}m{0.15\linewidth}
                    >{\centering\arraybackslash}m{0.15\linewidth}
                    >{\centering\arraybackslash}m{0.15\linewidth}
                    >{\centering\arraybackslash}m{0.15\linewidth}
                    >{\centering\arraybackslash}m{0.15\linewidth}}
    \scriptsize{Without sphere initialization} &
    \multicolumn{6}{m{0.9\linewidth}}{\includegraphics[width=\linewidth]{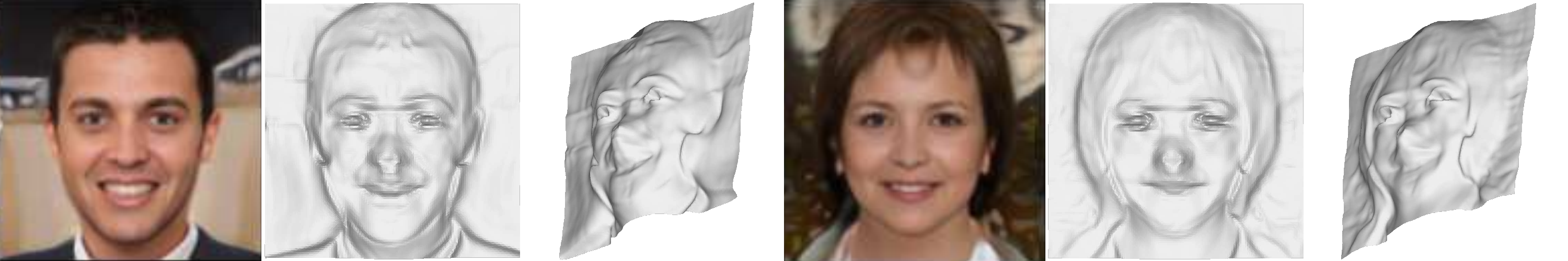}}\tabularnewline
    \scriptsize{With sphere initialization} &
    \multicolumn{6}{m{0.9\linewidth}}{\includegraphics[width=\linewidth]{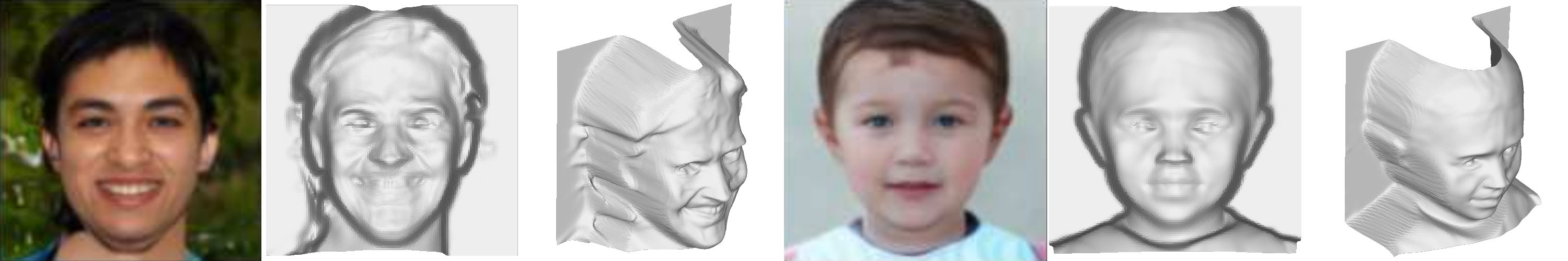}}\tabularnewline
    & \scriptsize{Color Image} & \scriptsize{Frontal view} & \scriptsize{Side view} & \scriptsize{Color Image} & \scriptsize{Frontal view} & \scriptsize{Side view}
    \end{tabular}
\caption{Sphere initialization ablation study. We visualize volume-rendered RGB and depth images from volume renderers trained with and without sphere initialization. The Depth map meshes are visualized from the front and side views. 
Note how a model trained without model sphere initialization generates concave surfaces.}
\label{fig:sphere_init_ablation}
\end{figure*}

\section{Sampling Strategy}
\label{sec:sampling_strategy}
NeRF~\cite{mildenhall2020nerf}, along with existing 3D-aware GANs like Pi-GAN~\cite{chan2021pi}, rely on hierarchical sampling strategy for obtaining more samples near the surface. 
Our use of SDFs allows sampling the volume with smaller number of samples without sacrificing the surface quality, thereby reducing the memory footprints and simplifying the implementation. 

Stratified sampling randomizes the distance between adjacent samples along each ray, adding undesired noise to the volume rendering (\cref{fig:stratified_ablation}). The randomness also amplifies flickering in RGB videos. Our  sampling strategy ensures that the integration intervals are of the same length, which eliminates the noise and results in smoother volume rendering outputs.

\section{Ablation studies}
\label{sec:ablations}
We perform two ablation studies to show the necessity of the minimal surface loss (see main paper) and the sphere initialization. As can be seen in~\Cref{fig:min_surf_ablation}, on top of preventing spurious and non-visible surfaces from being formed, the minimal surface loss also helps to disambiguate between shape and radiance. Penalizing values that are close to zero essentially minimizes the surface area and makes the network prefer smooth SDFs.

In~\Cref{fig:sphere_init_ablation}, we show the importance of the sphere initialization in breaking the concave/convex ambiguity. Without properly initializing the weights, the network gets stuck at a local minimum that generates concave surfaces. Although concave surfaces are physically incorrect, they can perfectly explain multi-view images, as they are essentially the "mirror" surface. Concave surfaces cause the images to be rendered in the opposite azimuth angle, an augmentation that the discriminator cannot detect as fake. Therefore, the generator cannot recover from the this local minima.

\section{Limitations (continued)}\label{sec: limitation}
As mentioned in the main paper, our high-resolution generation network is based on the implementation of StyleGAN2 \cite{karras2019style}, and thus might experience the same aliasing and flickering at regions with high-frequency details (e.g., hair), which are recently addressed in Alias-free GAN \cite{karras2021alias} or Mip-NeRF \cite{Barron_2021_ICCV}. 
Moreover, we observe that the reconstructed geometry for human eyes contain artifacts, characterized by concave, instead of convex, eye balls. We believe that these artifacts often lead to slight gaze changes along with the camera views. 
As stated in the main paper, our current implementation of volume rendering during inference uses fixed frontal view directions for RGB queries $\mathbf{c}(\mathbf{x},\mathbf{v})$, and thus cannot express moving specular highlights along with the camera. 

\section{Additional Results}
We show uncurated set of images generated by our networks (Fig.~\ref{fig:supp_additional}).
\begin{figure*}[th!] 
\includegraphics[width=1\linewidth]{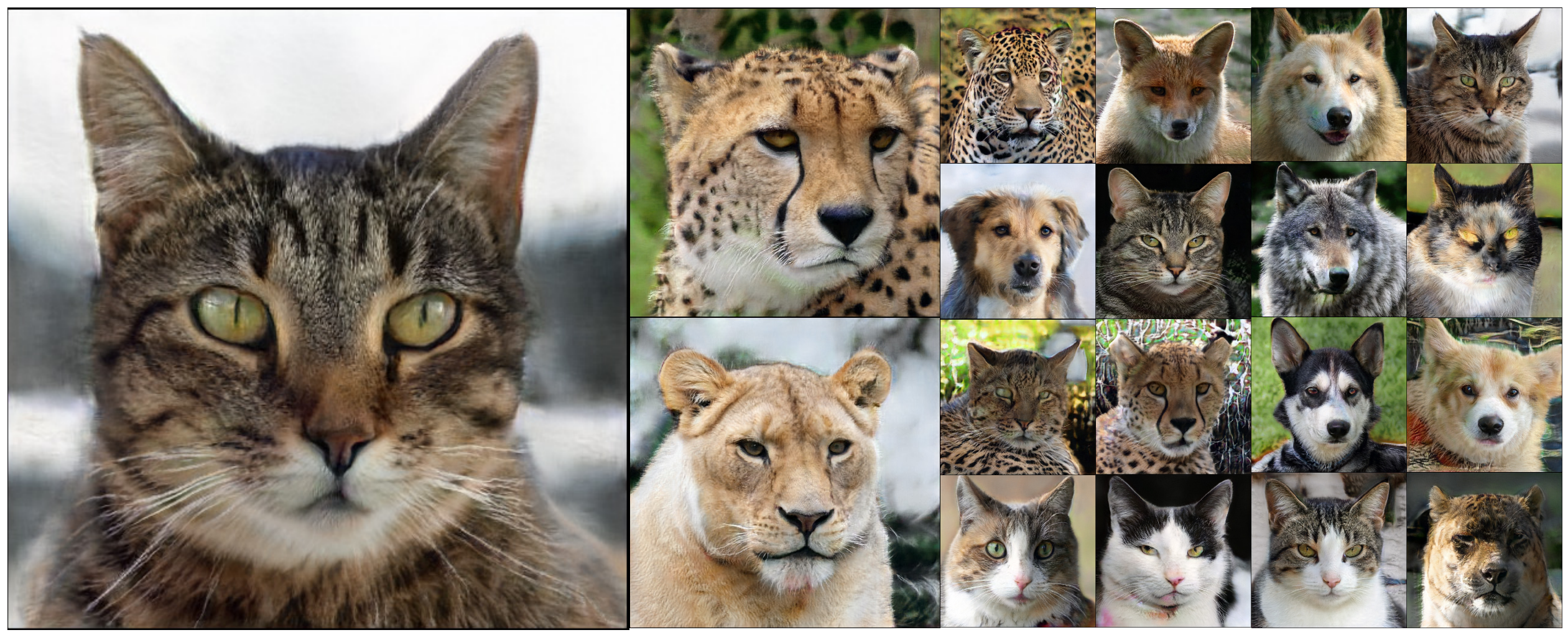}
\includegraphics[width=1\linewidth]{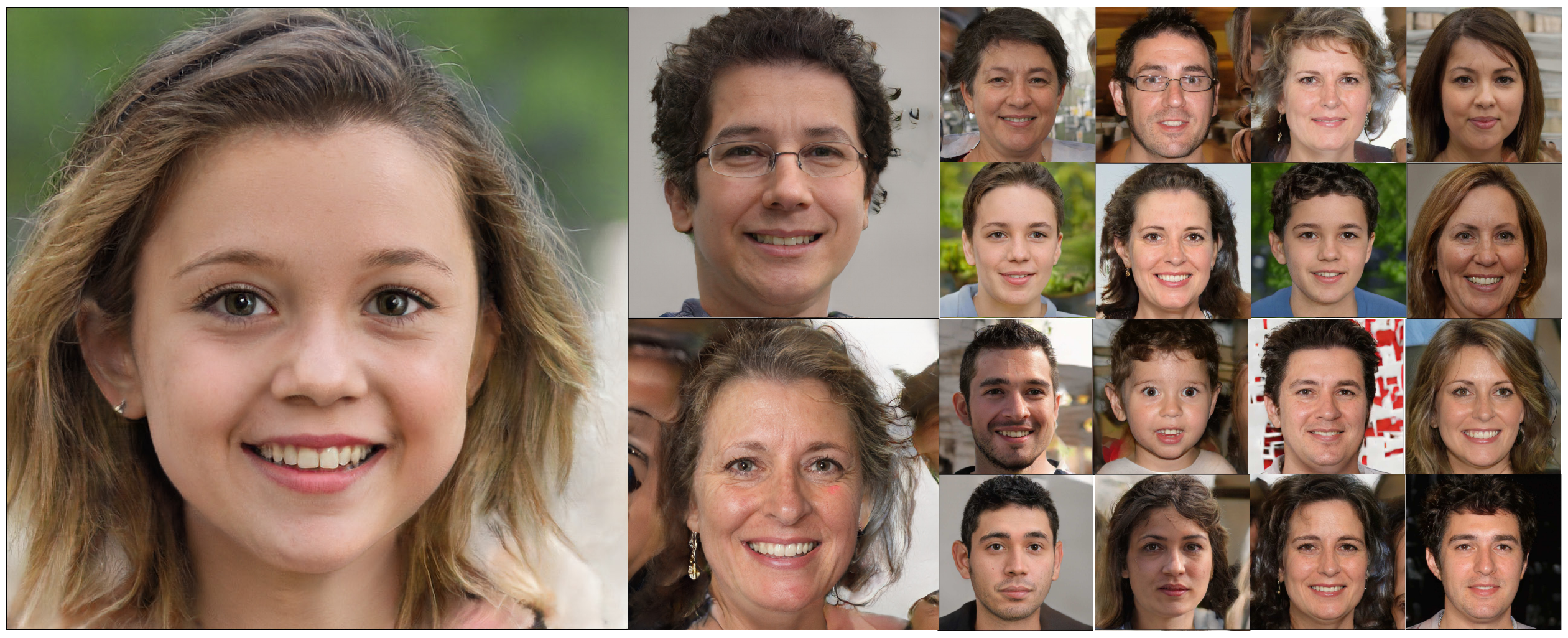}
\caption{Uncurated high-resolution RGB images that are randomly generated by StyleSDF.}\label{fig:supp_additional}
\end{figure*}

\fi

\end{document}